\documentclass[letterpaper]{article} 
\usepackage{aaai2026}  
\usepackage{times}  
\usepackage{helvet}  
\usepackage{courier}  
\usepackage[hyphens]{url}  
\usepackage{graphicx} 
\usepackage{natbib}  
\usepackage{caption} 
\usepackage{algorithm}
\usepackage{algorithmic}

\usepackage{newfloat}
\usepackage{listings}
\DeclareCaptionStyle{ruled}{labelfont=normalfont,labelsep=colon,strut=off} 
\floatstyle{ruled}
\newfloat{listing}{tb}{lst}{}
\floatname{listing}{Listing}
\usepackage{bm}
\usepackage{amsmath}
\usepackage{multirow}
\usepackage{amssymb}

\usepackage{booktabs}
\usepackage{makecell} 
\usepackage{float}    
\usepackage{array}
\usepackage{xcolor}  
\usepackage{colortbl} 

\usepackage{tcolorbox}   
\tcbuselibrary{skins}
\usepackage{placeins}
\usepackage{subcaption}

\definecolor{highlight}{RGB}{230, 244, 255}
\definecolor{diffplus}{RGB}{0, 127, 0}
\definecolor{diffminus}{RGB}{191, 0, 0}

\title{Not All Tokens Are What You Need In Thinking}
\author{
    Hang Yuan\textsuperscript{\rm 1,5},
    Bin Yu\textsuperscript{\rm 2,5},
    Haotian Li\textsuperscript{\rm 2,6},
    Shijun Yang\textsuperscript{\rm 3,6}, \\
    Christina Dan Wang\textsuperscript{\rm 4},
    Zhou Yu\textsuperscript{\rm 1},
    Xueyin Xu\textsuperscript{\rm 5,6},
    Weizhen Qi\textsuperscript{\rm 5,6},
    Kai Chen\textsuperscript{\rm 5,6}
}
\affiliations{
    \textsuperscript{\rm 1}East China Normal University,
    \textsuperscript{\rm 2}Harbin Institute of Technology,\\
    \textsuperscript{\rm 3}University of Science and Technology of China,
    \textsuperscript{\rm 4}New York University Shanghai,\\
    \textsuperscript{\rm 5}Zhongguancun Academy,
    \textsuperscript{\rm 6}Zhongguancun Institute of Artificial Intelligence\\
    Correspondence: weizhenqi@zgci.ac.cn
}

\begin{document}
\nocopyright

\maketitle

\begin{abstract}
Modern reasoning models, such as OpenAI's o1 and DeepSeek-R1, exhibit impressive problem-solving capabilities but suffer from critical inefficiencies: high inference latency, excessive computational resource consumption, and a tendency toward overthinking--generating verbose chains of thought (CoT) laden with redundant tokens that contribute minimally to the final answer.
To address these issues, we propose Conditional Token Selection (CTS), a token-level compression framework with a flexible and variable compression ratio that identifies and preserves only the most essential tokens in CoT.
Using conditional importance scoring, CTS evaluates each token's contribution to deriving correct answers and then trains models on compressed CoT. 
Extensive experiments demonstrate that CTS effectively compresses long CoT while maintaining strong reasoning performance.
Notably, on the GPQA benchmark, Qwen2.5-14B-Instruct trained with CTS achieves a 9.1\% accuracy improvement with 13.2\% fewer reasoning tokens (13\% training token reduction). Further reducing training tokens by 42\% incurs only a marginal 5\% accuracy drop while yielding a 75.8\% reduction in reasoning tokens, highlighting the prevalence of redundancy in existing CoT.
\end{abstract}


\section{Introduction}

Large reasoning models such as o1\citep{jaech2024openai} and R1\citep{guo2025deepseek} significantly enhance their reasoning capabilities through reinforcement learning, instructing models to generate thoughtful reasoning steps before producing final answers. \citet{guo2025deepseek} demonstrated that by fine-tuning non-reasoning models like Qwen2.5-14B-Instruct on long Chain of Thought (CoT) data generated by R1, these models can acquire comparable reasoning abilities, even surpassing o1-mini on math and code reasoning tasks.
Consequently, numerous distilled R1 reasoning datasets have emerged, including s1K\cite{muennighoff2025s1simpletesttimescaling}, SkyThought\citep{sky_t1_2025}, OpenMathReasoning\citep{moshkov2025aimo}, and AM-1.4M \citep{zhao202514millionopensourcedistilled}. 
Small-scale language models trained on these datasets consistently demonstrate remarkable reasoning capabilities.
However, the ever-increasing length of CoT sequences burdens both training and inference, with recent studies \citep{sui2025stop} revealing that models often overthink, expending substantial resources on redundant reasoning steps.

\begin{figure}
    \centering
    \includegraphics[width=0.91\linewidth]{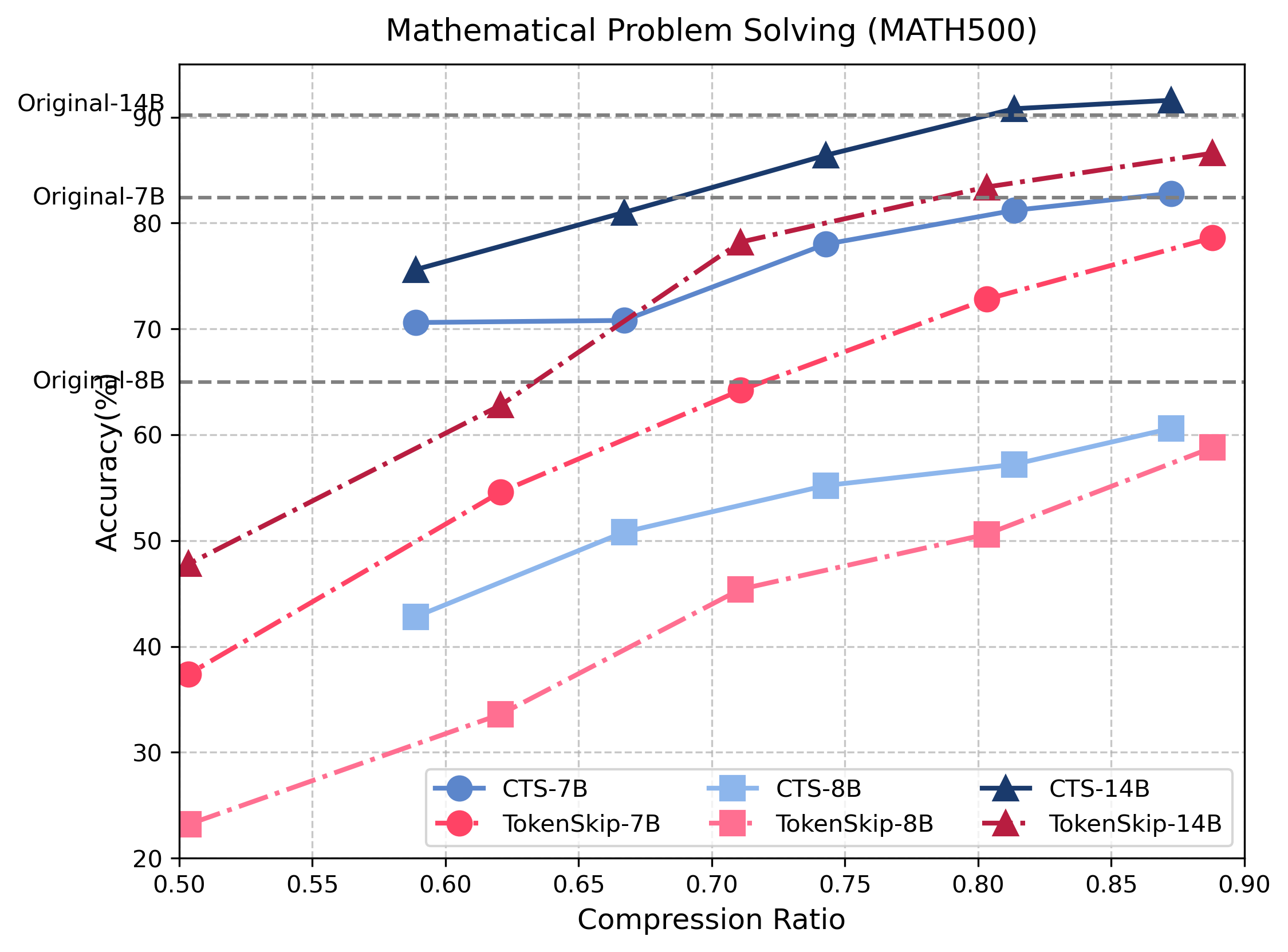}
    \caption{Comparison of accuracy across different compression ratios for various model configurations.}
    \label{fig:math500_acc}
\end{figure}

This inefficiency raises a critical question: "\textbf{How can we maintain the high performance of long CoT reasoning while eliminating its computational waste?}" Existing solutions, such as TokenSkip’s task-agnostic compression \citep{xia2025tokenskip}, show promise for short CoT sequences but fail to address the unique challenges of reinforcement learning-generated long CoT data (spanning thousands of tokens).
Moreover, they focus solely on compressing the CoT trajectory while overlooking contextual signals like questions and answers, which prior work \citet{tian2025think} identifies as key to effective compression.

\begin{figure*}
    \centering
    \includegraphics[width=\textwidth]{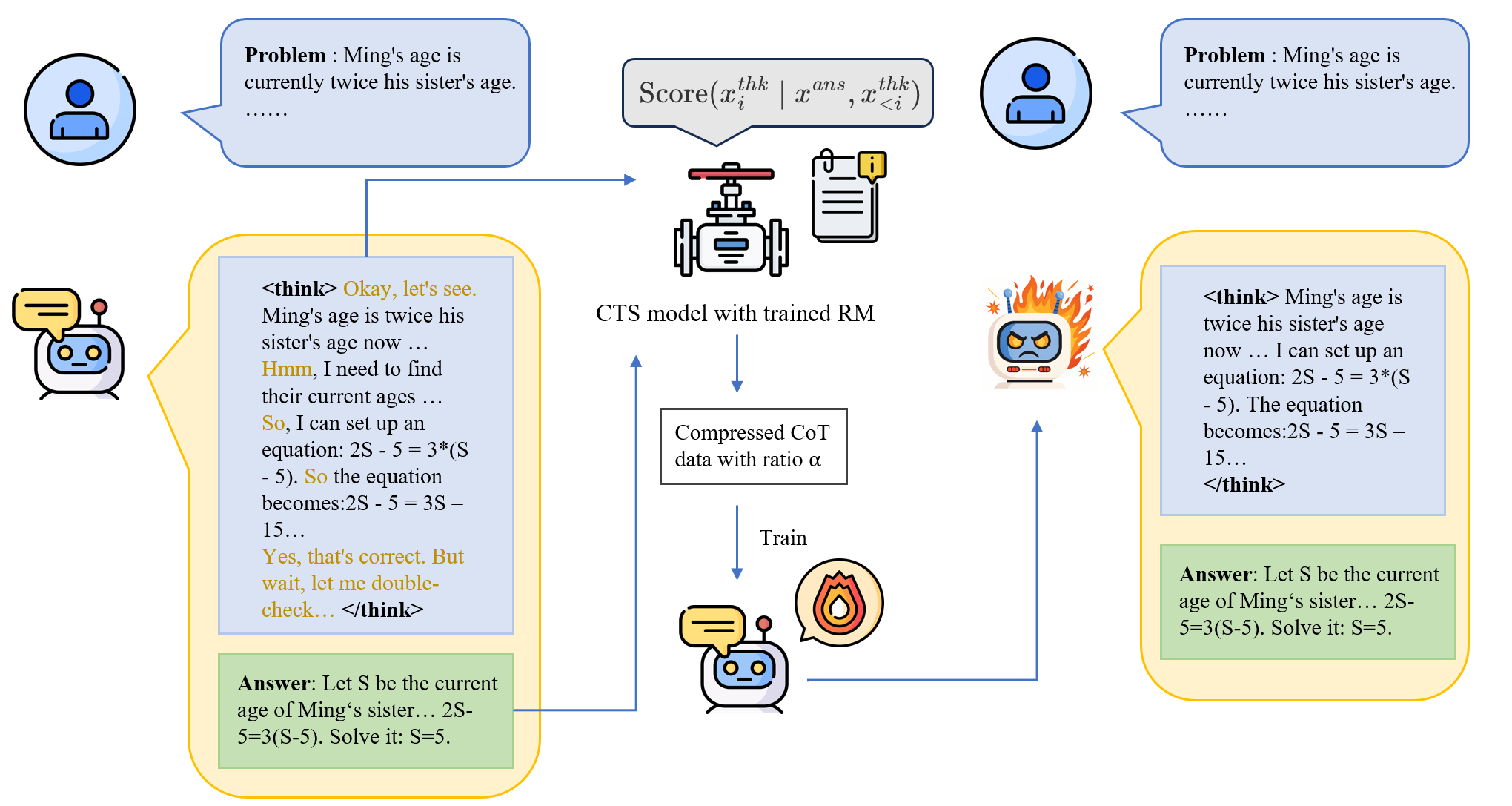}
    \caption{
    Illustration of Conditional Token Selection. For long CoT datasets, CTS leverages a well-trained reference model to evaluate the importance of each thinking token conditional on the answer, removing less important tokens based on the compression ratio $\alpha$. The model is then trained on this compressed data, enabling more efficient reasoning capabilities.
    }
    \label{fig:process}
\end{figure*}

To bridge this gap, we propose Conditional Token Selection, a framework that dynamically prunes redundant reasoning tokens while preserving those essential for deriving answers.
CTS leverages a reference model (RM) trained on high-quality reasoning corpora to score token importance conditioned on critical context (e.g., questions and answers). 
As shown in Figure \ref{fig:process}, by filtering CoT data at adjustable compression ratios and then fine-tuning model with compressed data, we enable models to learn how to skip unnecessary reasoning tokens during inference.

We conducted extensive experiments on models of various sizes, including the Llama-3.1-8B-Instruct \citep{grattafiori2024llama3herdmodels} and the Qwen2.5-Instruct series \citep{yang2024qwen2}.
The experimental results demonstrate the effectiveness of our method and confirm that there indeed exist many redundant thinking tokens in long CoT data.
As illustrated in Figure \ref{fig:math500_acc}, our approach demonstrates comprehensive accuracy improvements over the prominent TokenSkip \citep{xia2025tokenskip} across all three models on the MATH500 benchmark \citep{hendrycks2021measuring}.
Additionally, when applied to Qwen2.5-14B-Instruct on the GPQA benchmark \citep{rein2024gpqa}, CTS demonstrates remarkable efficiency gains: a modest 13\% reduction in training tokens results in both a \textbf{9.1\%} accuracy improvement and 13.2\% fewer reasoning tokens. When training token reduction is increased to \textbf{42\%}, the method incurs only a minor 5\% accuracy decrease while achieving a dramatic \textbf{75.8\%} reduction in reasoning tokens, highlighting an advantageous accuracy-efficiency trade-off.
Furthermore, although the model was trained on mathematical datasets, our experiments demonstrate its effectiveness on out-of-domain tasks, including code generation benchmarks such as HumanEval \cite{chen2021evaluating} and MBPP \cite{austin2021program}.

In summary, our key contributions are:
\begin{itemize}
    \item We introduce the Conditional Token Selection framework, which dynamically assigns importance scores to tokens in CoT trajectories, enabling flexible and context-aware preservation of critical reasoning tokens.
    
    \item  We've developed a reference model, trained on high-quality reasoning data, that more accurately assesses the conditional importance of tokens in reasoning CoTs. This model can be applied to other independent tasks, such as prompt compression.
    
    \item We comprehensively compare token-based compression methods both conditional and non-conditional for long CoT data generated by reinforcement learning, thereby validating our token selection strategies.
    
\end{itemize}

\section{Preliminaries}

In this section, we introduce some important preliminary concepts. Unless specified, boldface lowercase letters (e.g., $\bm{x}, \bm{y}$) are used to represent concatenated token sequences, whereas individual tokens are denoted by subscripted lowercase letters (e.g., $x_i, y_i$). $\bm{x}_{<t}$ represents the subsequence of $\bm{x}$ containing all tokens with indices less than $t$.


For a given context $\bm{x}=\{x_i\}_{i=1}^{n}$, the self-information of a token $x_t$ ($1\leq t \leq n$) can be defined as:
\begin{equation}\label{equ:self-info}
    \begin{split}
        I(x_t) &= -\log_2P(x_t \mid \bm{x}_{<t}) \\
        &=-\log_2P(x_t \mid x_1,...,x_{t-1})
    \end{split}
\end{equation}
Perplexity (PPL) is then defined based on self-information as:
\begin{equation}\label{equ:perplexity}
    \text{PPL}(x_t)=2^{I(x_t)}
\end{equation}
Perplexity is commonly used to measure a language model's ability to predict the given context. 
Since removing tokens with lower perplexity \citep{li2023compressing} has a relatively minor impact on the model's understanding and prediction of the context, perplexity can serve as a viable metric for evaluating token importance.
To mitigate information redundancy in prompts with long contexts and avoid resource waste (e.g., reducing tokens consumption in LLM API calls), \citet{li2023compressing} proposed Selective Context that uses a small language model to assess the perplexity of each lexical unit (such as sentences or tokens) in the original prompts, and then drops the less informative content for prompt compression.

Selective context is an unconditional approach, as it does not utilize existing important information (such as the standard answers to math problems, which could serve as key references for simplifying or compressing lengthy reasoning steps) when calculating token importance. Next, we introduce conditional perplexity as a more fine-grained metric for token importance, which will be employed in our proposed method.
Given essential information $\bm{v}$, the conditional self-information and perplexity can be defined as: 
\begin{equation}
    \begin{split}
        &I(x_t \mid \bm{v}) = -\log_2 P(x_t \mid \bm{v}, \bm{x}_{<t}) \\
        &\text{PPL}(x_t\mid \bm{v}) = 2^{I(x_t\mid \bm{v})}
    \end{split}
\end{equation}

\section{Methodology}
Building upon the concept of utilizing essential information as conditions, we propose \textbf{C}onditional \textbf{T}oken \textbf{S}election (CTS). 
It employs a well-trained reference model to evaluate conditional perplexity for compressing long CoT sequences, such as those generated by harnessing reinforcement learning techniques (e.g., R1) and subsequently fine-tunes models using this compressed data.
While small models fine-tuned on original long CoT data exhibit strong reasoning capabilities, they suffer from slow inference speed, redundant reasoning processes, and high computational resource consumption.
By selectively retaining critical tokens, CTS aims to enhance model performance while significantly reducing both training and inference resource overhead.

\begin{figure}
    \centering
    \includegraphics[width=\linewidth]{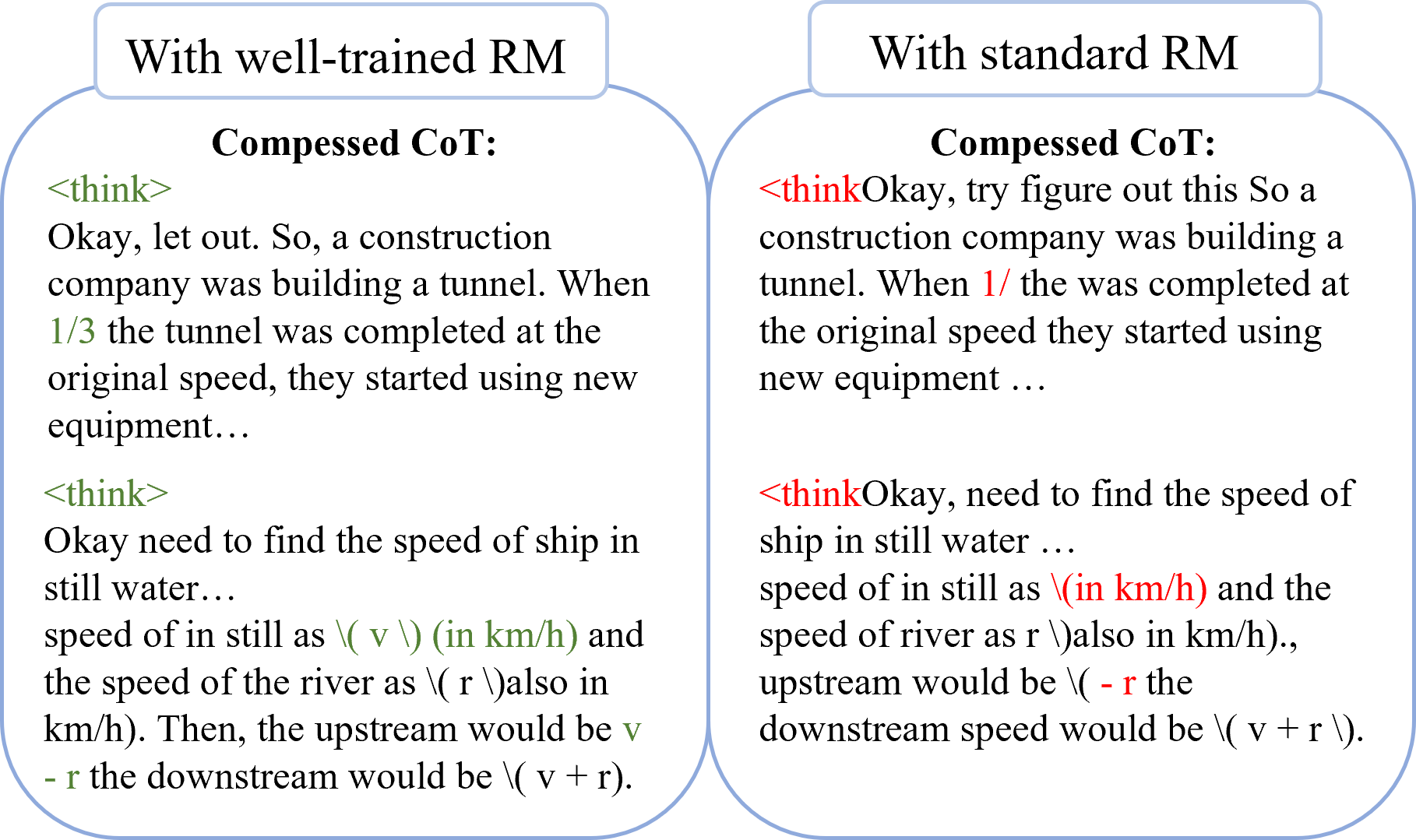}
    \caption{An example of CoT compression using CTS, where the left figure employs a well-trained RM, while the right uses a standard RM.}
    \label{fig:withRM}
\end{figure}

\subsection{Problem Formulation}
Given a long CoT dataset, each instance $\bm{x}$ is a token sequence composed of three segments: a problem statement $\bm{x}^{\text{prob}}$, a chain-of-thought reasoning process $\bm{x}^{\text{thk}}$, and a final answer $\bm{x}^{\text{ans}}$. Formally, $\bm{x}$ is structured as the concatenation $\bm{x} = \{\bm{x}^{\text{prob}}, \bm{x}^{\text{thk}}, \bm{x}^{\text{ans}}\}$, illustrated in Figure \ref{fig:process}. 

In the process of distillation, the Large Reasoning Model (e.g. R1 and o1) initially generates
long CoT data, which is subsequently employed to fine-tune small-scale language models (e.g., Qwen2.5-14B-Instruct). The primary objective of distillation approach is to endow smaller language models with enhanced reasoning capabilities. Then these distilled models can achieve strong performance under resource-constrained conditions. The loss function of distillation can be reformulated as:
\begin{equation}
    \mathcal{L} = -\sum_{i=1}^{l} \log P_{\theta}(y_i \mid \bm{x}^{\text{prob}}, \bm{y}_{<i})
\end{equation}
where $\bm{y}=\{y_i\}_{i=1}^l = \{\bm{x}^{\text{thk}}, \bm{x}^{\text{ans}}\}$ is the target sequence of $l$ tokens and $\theta$ are model parameters.
However, the ever-increasing length of CoT sequences burdens both training and inference.

Conditional Token Selection identifies and retains critical reasoning tokens in long CoT training data while eliminating redundant tokens that hinder model reasoning efficiency.
When fine-tuned on CTS-filtered CoT data, the model retains strong reasoning capabilities while significantly reducing inference token consumption, thereby enhancing reasoning efficiency. 
The objective of CTS can be defined as:
\begin{equation}
    \sideset{}{}{\arg\min}_{\tilde{\bm{y}}} \left\{\text{dist}(A, \tilde{A}) + \lambda\|\tilde{\bm{y}}\|_0\right\},
\end{equation}
Where $\tilde{\bm{y}}$ represents the compressed CoT, a subsequence of $\bm{y}$.
$A$ and $\tilde{A}$ represent the answers to any question $Q$ given by the small language models trained with $\bm{y}$ and $\tilde{\bm{y}}$, respectively.
Here, $\text{dist}(\cdot,\cdot)$ is a function measuring the distance (e.g., KL divergence). $\lambda$ serves as a hyper-parameter balancing the compression ratio and distance.
The $\|\cdot\|_0$ norm serves as a penalty term to control the complexity of the compressed data $\tilde{\bm{y}}$, preventing trivial compression (i.e. $\tilde{\bm{y}} = \bm{y}$). This regularization term is also incorporated in Equation 1 \citep{jiang2023longllmlingua}.

\subsection{Reference Modeling}

When compressing CoT data, a \textbf{R}eference \textbf{M}odel (RM) is typically required to estimate self-information or perplexity using Equations \eqref{equ:self-info} and \eqref{equ:perplexity}. This RM is usually a lightweight small language model (e.g. Qwen2.5-7B-Instruct, Llama3-8B-Instruct).
However, through experimental observations, we found that using RM directly to compress CoT tends to remove important but commonly used numbers or alphanumeric symbols. 
As depicted in the right panel of Figure \ref{fig:withRM}, both the water flow velocity variable '$v$' and the denominator '3' from the fraction 1/3 have been omitted. These elements are essential for comprehending the subsequent equations.

To teach the RM which variables and logical symbols are critical for deriving the final answer, we can modify the model's conditional output distribution by increasing the probabilities of symbols when reasoning steps are used as conditions.
First, we collected mathematical reasoning data containing long CoT sequences.
These were initially filtered using large language models with strong mathematical capabilities (e.g., DeepSeek-R1) to retain only the most crucial reasoning steps, with particular attention to preserving critical numerical values and logical symbols. Subsequently, we conducted manual screening to eliminate problems with incorrect reasoning outcomes (see Appendix for more details).
It is worth noting that this portion of CoT data differs from the distilled data requiring compression in our method.

Each instance in the curated dataset that preserves critical variables and logical symbols can be represented as a token sequence $\{\bm{q},z_1,\ldots,z_K,\bm{a}\}$ where, $q$ denotes the question, $a$ represents the answer, and $z_1,\ldots,z_K$ are the reasoning steps retaining essential symbolic information.
The RM can be trained using the following cross-entropy loss:
\begin{equation}
    \mathcal{L}_{RM}=-\sum_{i=1}^K \log P(z_i\mid x^{ins}, \bm{q},\bm{a}, \bm{z}_{<i})
\end{equation}
Where $x^{ins}$ represents "\textit{For a problem $\bm{q}$, the following reasoning steps are important to get the answer $\bm{a}$}".  
The resulting RM is then used to assess the token importance within long CoT trajectories, allowing us to focus on the most influential tokens.

\subsection{Token Selection Based on Conditional Perplexity Differences}
To reduce token consumption during training while maintaining powerful and efficient reasoning capabilities of small-scale language model, we apply conditional compression to CoT data.
We first employ the RM to compute conditional perplexity scores for each token in target sequences $\bm{y}$, given question and answer $\bm{x}^{\text{prob}},\bm{x}^{\text{ans}}$.
Then token compression with a flexible ratio can be achieved by sorting all tokens in the context by perplexity and subsequently removing lower-ranking tokens.

When utilizing perplexity for token compression, to ensure the compressed data and original data distributions are the same, the following conditional independence assumption is required:
\begin{equation}\label{indep}
    A \perp \!\!\! \perp  \bm{\bar{y}} \mid \bm{\tilde{y}}
\end{equation}
Here, $\bm{\bar{y}}$ denotes the removed tokens from the target sequence $\bm{y}$, which is composed as the concatenation of retained tokens $\bm{\tilde{y}}$ and removed tokens $\bm{\bar{y}}$, i.e., $\bm{y}=\bm{\tilde{y}} \oplus \bm{\bar{y}}$.
Under this assumption, the model can derive correct answer from the compressed reasoning data $\bm{\tilde{y}}$, satisfying:
\begin{equation}
    p(A\mid \bm{\tilde{y}}) \approx p(A\mid \bm{y})
\end{equation}

To mitigate the inaccuracy caused by the conditional independence assumption, we adopt the iterative compression approach proposed by \citet{jiang2023llmlingua}. This method comprises three key steps:(1) uniformly partitioning $\bm{y}$ into $m$ segments $\mathcal{S}=\{\bm{s}_1,\ldots,\bm{s}_{m}\}$; (2) applying CTS to each segment to generate $\{\tilde{\bm{s}}_1,\ldots,\tilde{\bm{s}}_{m}\}$; (3) concatenating the compressed segments to produce the final compressed sequence $\bm{\tilde{y}}$.

For more fine-grained assessment of token conditional importance, instead of using conditional perplexity scores directly, we use the distribution shift caused by the condition of the answer to quantify the association between the thinking token and the answer. Consequently, we derive a score, $r_i$, for each token in the target sequence $y$, calculated as follows:
\begin{equation*}
    r_i = \text{PPL}(y_i\mid \bm{y}_{<i}) - \text{PPL}(y_i\mid \bm{x}^{\text{prob}}, \bm{x}^{\text{ans}}, \bm{y}_{<i})
\end{equation*}

As shown in Alg. \ref{alg:algorithm}, given a compression ratio $\alpha$, after scoring each token in target sequence using the RM, we determine a threshold $r_{\alpha}$ represented by the $\alpha$-quantile of importance scores, thereby selecting reasoning tokens whose scores exceed the threshold:
$$ \tilde{\bm{y}} = \{ y_i \mid r_i > r_{\alpha} \}.$$

\begin{algorithm}[tb]
\caption{CoT Compression Algorithm}
\label{alg:algorithm}
\textbf{Input}: Long CoT data $\bm{x}$, compression ratio $\alpha \in (0,1)$, target sequence $\bm{y}$\\
\textbf{Output}: Compressed sequence $\tilde{\bm{y}}$
\begin{algorithmic}[1]
\STATE Set number of partitions $m$
\STATE Initialize compressed sequence $\tilde{\bm{y}} = [\,]$
\STATE Uniformly partition $\bm{y}$ into $m$ sub-sequences: $\mathcal{S} = \{s_1, s_2, \ldots, s_m\}$

\FOR{$i = 1$ \TO $m$}
    \STATE $L_i \gets |s_i|$ \hfill // sub-sequence length
    \STATE $k_i \gets \lceil \alpha \cdot L_i \rceil$ \hfill // tokens to keep
    
    \FOR{$j = 1$ \TO $L_i$}
        \STATE Compute distribution shift score $r_{i,j}$ for token $s_{i,j}$ using RM
    \ENDFOR
    
    \STATE $\mathcal{J}_i \gets$ indices of top-$k_i$ tokens by score $r_{i,j}$
    \STATE Sort $\mathcal{J}_i$ in ascending order to preserve original sequence
    \STATE $\tilde{s}_i \gets [s_{i,j}]_{j \in \mathcal{J}_i}$ \hfill // compressed sub-sequence
    \STATE $\tilde{\bm{y}} \gets \tilde{\bm{y}} \oplus \tilde{s}_i$ \hfill // concatenation
\ENDFOR

\STATE \textbf{return} $\tilde{\bm{y}}$
\end{algorithmic}
\end{algorithm}

\section{Experiments}

\begin{table*}[htbp]
  \centering
    \caption{Experimental results of various compression methods on \textbf{Qwen2.5-14B-Instruct}, showing accuracy, average reasoning CoT tokens, and compression ratio (actual ratio).}
  \resizebox{\textwidth}{!}{
  \begin{tabular}{@{}lccccccc@{}}
    \toprule
    \multirow{2}{*}{Methods} & \multirow{2}{*}{Ratio (Actual)} & \multicolumn{2}{c}{MATH500} & \multicolumn{2}{c}{AIME24} & \multicolumn{2}{c}{GPQA Diamond} \\
    \cmidrule(lr){3-4} \cmidrule(lr){5-6} \cmidrule(lr){7-8}
    & & Accuracy $\uparrow$ & Tokens $\downarrow$ & Accuracy $\uparrow$ & Tokens $\downarrow$ & Accuracy $\uparrow$ & Tokens $\downarrow$ \\
    \midrule
    Original & 1.0 & $ 90.2$ & $5012$ & $40$ & $23041$ & $51.5$ & 12000 \\
    \midrule
    \multirow{1}{*}{GPT-4o} & 0.9(0.06) & $61.4$ & $\textbf{283}$ & $0$ & $\textbf{353}$ & $35.8$ & $\textbf{353}$ \\
    \midrule
    \multirow{5}{*}{LLMLingua} 
    & 0.9(0.89) & $84.4$ & $5023$ & $33.3$ & $19731$ & 53.0 & 10689 \\
    & 0.8(0.81) & $65.6$ & $2510$ & $10.0$ & $4230$ & 44.9 & 4037 \\
    & 0.7(0.73) & $60.4$ & $2511$ & $6.7$ & $4588$ & 43.9 & 3371 \\
    & 0.6(0.62) & $59.0$ & $2270$ & $6.7$ & $3076$ & 40.9 & 3347 \\
    & 0.5(0.55) & $53.6$ & $1998$ & $3.3$ & $3789$ & 40.4 & 2796 \\
    \midrule
    \multirow{5}{*}{TokenSkip} 
    & 0.9(0.88) & $86.6$ & $4941$ & $40.0$ & $19985$ & 50.0 & 12455 \\
    & 0.8(0.80) & $83.4$ & $5549$ & $26.7$ & $20945$ & 50.0 & 13275 \\
    & 0.7(0.71) & $78.2$ & $6566$ & $16.7$ & $24718$ & 43.4 & 15531 \\
    & 0.6(0.62) & $62.8$ & $8595$ & $10.0$ & $27748$ & 38.8 & 16764 \\
    & 0.5(0.50) & $47.8$ & $9824$ & $3.3$ & $26555$ & 31.8 & 19121 \\
    \midrule
    \multirow{5}{*}{\textbf{CTS}}
    & 0.9(0.87) & $\textbf{91.6}$ & $4703$ & $\textbf{50.0}$ & $21668$ & \textbf{60.6} & 10413 \\
    & 0.8(0.81) & $\textbf{90.8}$ & $4922$ & $\textbf{43.4}$ & $21767$ & \textbf{53.5} & 13136 \\
    & 0.7(0.74) & $\textbf{86.4}$ & $3310$ & $\textbf{33.3}$ & $10448$ & \textbf{57.1} & 10372 \\
    & 0.6(0.66) & $\textbf{81.0}$ & $3787$ & $\textbf{16.7}$ & $10308$ & \textbf{48} & 9712 \\
    & 0.5(0.58) & $\textbf{75.6}$ & $2036$ & $\textbf{10.0}$ & $3196$ & \textbf{46.5} & 2906 \\
    
    \bottomrule
  \end{tabular}
  }
  \label{tab:14B_all} 
\end{table*}

\subsection{Experimental Setup}

\subsubsection{Reference Model Training}
To train our mathematical reasoning reference model, we utilized the first 9.3K samples from the MATH training dataset \citep{openr1}, comprising problems from NuminaMath 1.5 with reasoning traces generated by DeepSeek-R1. To imbue the model with the ability to assess token importance in mathematical reasoning chains, DeepSeek-R1 was prompted to meticulously select 8 million tokens from an original pool of 54 million reasoning tokens. This curated set of tokens is subsequently employed to supervise the fine-tuning of the RM.

\subsubsection{Implementation Details \& Dataset} We adopt the second 9.3K subset of MATH \citep{openr1} as the original long CoT data to be compressed. Building upon the framework introduced by \citet{jiang2023longllmlingua}, we perform conditional token selection using our
trained RM. We then fine-tuned Llama-3.1-8B-Instruct \citep{grattafiori2024llama3herdmodels}, Qwen2.5-7B/14B-Instruct \citep{yang2024qwen2} using the compressed dataset with varying compression ratios $\alpha \in \{0.5, 0.6, 0.7, 0.8, 0.9\}$. 

\subsubsection{Evaluation Benchmarks \& Metrics} The evaluation leverages three widely used reasoning benchmarks: \textbf{AIME24}, \textbf{MATH500}, and \textbf{GPQA Diamond} \cite{AoPS_AIME_2024, hendrycks2021measuring, rein2024gpqa}. 
We used the actual compression ratio, average accuracy, and average reasoning token count as metrics to evaluate compression methods.
All training and evaluation were conducted on 8 NVIDIA A800 GPUs. During training, we fine-tuned for 3 epochs with a batch size of 16. The maximum learning rate was set at 1e-5 with a cosine decay schedule. We set the maximum sequence length to 4096.

\subsubsection{Baselines} In our main experiments, we compared Conditional Token Selection with \textbf{unconditional TokenSkip} \citep{xia2025tokenskip}, \textbf{LLMLingua} \citep{jiang2023llmlingua} and \textbf{Prompt-based Compression}. We designate the method that directly uses the original CoT data for distillation as \textbf{Original}. For the Prompt-based method, we instructed GPT-4o\footnote{We use the gpt-4o-2024-08-06 version for experiments.} to compress long CoT trajectories by providing prompts such as "Please retain important reasoning tokens in the Chain-of-Thought and remove unnecessary ones, preserving $\alpha$\% of the original tokens." However, we observed that GPT tends to be overly aggressive in compression, consistently preserving less than 10\% of the original tokens regardless of the specified $\alpha$ value. Therefore, we did not focus on the compression ratio $\alpha$ and simply used GPT-4o to compress the reasoning chains directly.
These baselines are referred to as \textbf{GPT-4o}, \textbf{LLMlingua} and \textbf{TokenSkip} in Table \ref{tab:14B_all}, respectively. 

\subsection{Main Results}

Table \ref{tab:14B_all} presents the performance of different compression methods on the Qwen2.5-14B-Instruct model across various compression ratios. 
Notably, our method achieves the highest accuracy across all five compression ratios compared to other compression approaches. Figure \ref{fig:math500_acc} illustrates the accuracy of CTS versus TokenSkip across different models and compression ratios. CTS consistently outperforms TokenSkip, highlighting the superiority of conditional token compression.

Although the Prompt-based GPT-4o shown in Table \ref{tab:14B_all} uses fewer inference tokens, this is a result of its excessive compression of the chain of thought. In reality, the model fails to learn thinking capabilities from long CoT data, resulting in low accuracy across all three benchmarks.

Furthermore, for CTS, when the compression ratio is 0.9 and 0.8, the model shows improvements in both accuracy and inference efficiency. For example, on AIME24, at compression ratios of 0.9 and 0.8, accuracy increased by 10\% and 4.3\% respectively, while inference tokens decreased by 1373 and 1274. Notably, for GPQA Diamond, CTS achieves a 9.1\% accuracy gain with 13.2\% fewer reasoning tokens. Further reduction of inference tokens by 75.8\% results in only a 5\% accuracy drop, and since the compression ratio at this point is 42\%, training costs are significantly reduced.

As the compression ratio decreases, although CTS accuracy declines compared to Original, inference tokens continue to decrease. Thus, there exists a trade-off between accuracy and inference efficiency. From Table \ref{tab:14B_all}, we can infer that the optimal point lies between ratios 0.7 and 0.8, where model capability remains unchanged while minimizing inference token consumption.

\begin{figure}
    \centering
    \includegraphics[width=0.9\linewidth]{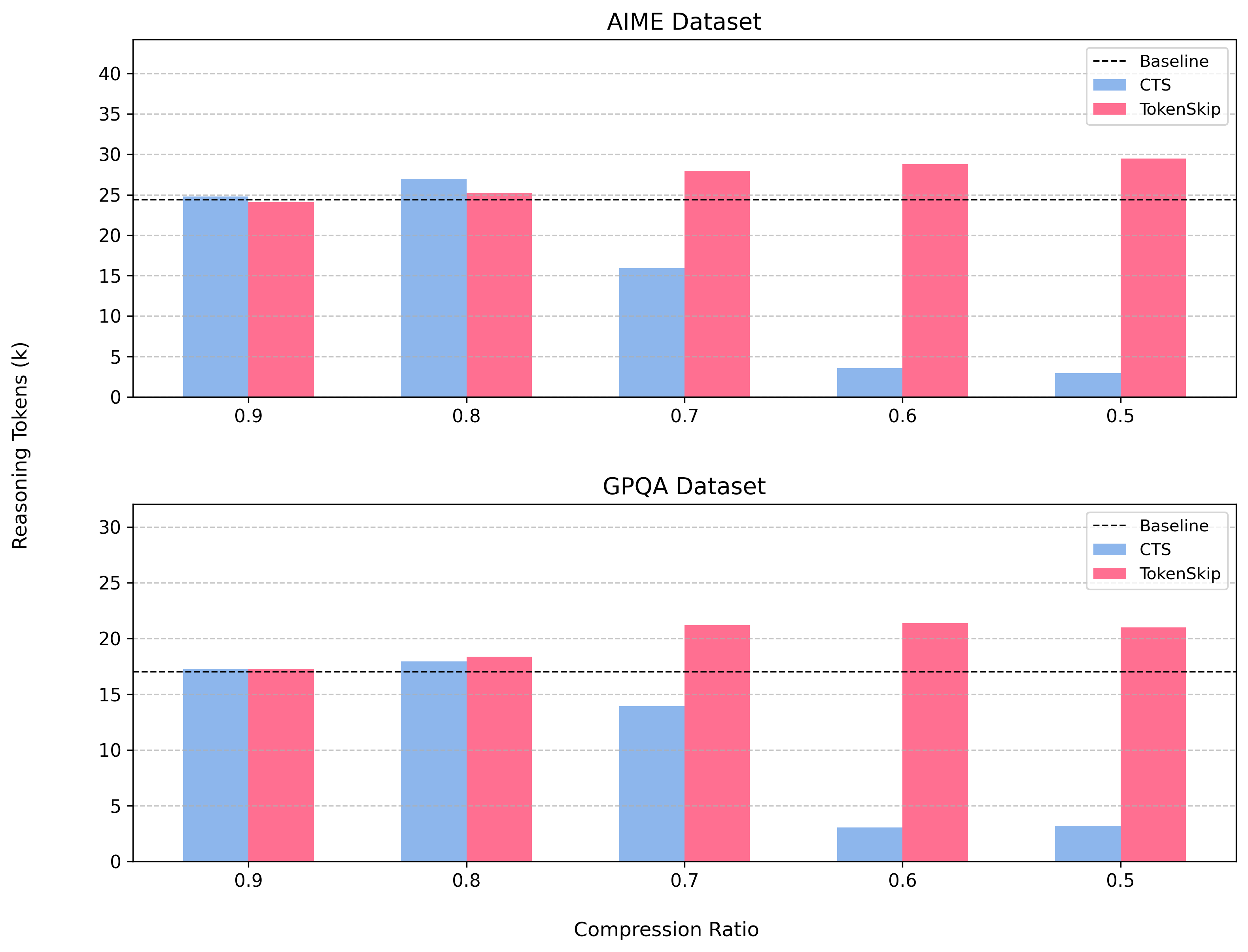}
    \caption{Compression Ratio vs. Average Reasoning Tokens for CTS and TokenSkip on the Qwen2.5-7B-Instruct model across different test sets.}
    \label{fig:all_tokens}
\end{figure}

However, poor compression methods can actually decrease model inference efficiency. As demonstrated in Figure \ref{fig:all_tokens}, TokenSkip's token consumption during reasoning exhibits an unexpected increase with higher compression ratios. 
Through meticulous examination of the model's reasoning process during testing, we observed that models trained on TokenSkip-compressed data tend to enter repetitive loops when addressing challenging queries, persistently outputting certain phrases until reaching the maximum token limit of 32,768. We speculate that as the compression ratio decreases, TokenSkip eliminates crucial reasoning-termination tokens, thereby impairing the model's reasoning capabilities. 
The primary reason is that TokenSkip's token importance metric was trained on MeetingBank data \citep{xia2025tokenskip}. While effective for compressing short CoT sequences, it fails to handle more challenging long CoT data containing complex reasoning steps.
A comprehensive discussion can be found in the Appendix.

Figure \ref{fig:acc_tokens} displays the percentage of reasoning token reduction and accuracy improvement for three methods. Higher accuracy with greater reasoning token reduction is preferable. Therefore, our method achieves the optimal balance between token reduction and accuracy improvement.

We've considered the reference model's overhead. Training the RM required 11.44 A800 GPU hours ($4.13\times 10^{18}$). For Qwen-7B-Instruct with CTS-14B at 0.7 compression ratio, inference saves $1.5\times 10^{14}$ FLOPs/inference, reaching breakeven at 27,500 calls. For MMLU Science tasks (16,000 questions) \citep{hendryckstest2021}, just two evaluations suffice, making the training cost negligible. (See Appendix)

\subsection{Ablation Study and Generalization Research}

\begin{figure}
    \centering
    \includegraphics[width=0.9\linewidth]{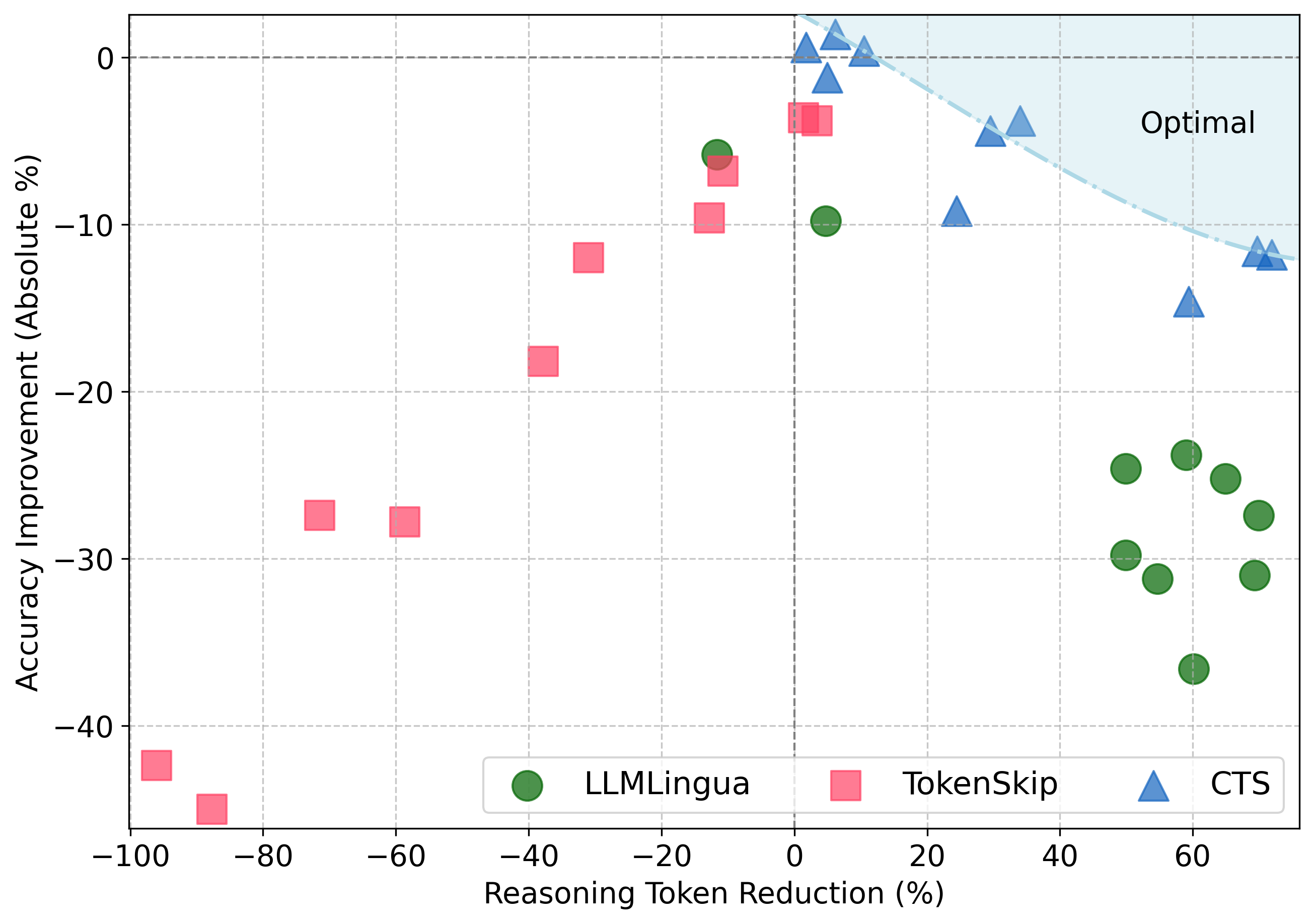}
    \caption{Performance metrics (Reasoning Token Reduction percentage and Absolute Accuracy Improvement (\%) relative to original) for various compression methods and ratios on Qwen2.5-7B-Instruct and Qwen2.5-14B-Instruct models and MATH500. The top-right region represents optimal performance, signifying higher accuracy and reduced reasoning token usage.}
    \label{fig:acc_tokens}
\end{figure}

To investigate the generalization capability of our approach, we evaluated CTS on two widely-used code datasets: HumanEval \cite{chen2021evaluating} and MBPP \cite{austin2021program}. 
Due to space constraints, we present only the results on the HumanEval benchmark in the main text, with comprehensive details provided in the Appendix.
As shown in Figure \ref{fig:generalization}, although trained solely on mathematical datasets, our model achieves higher accuracy and fewer reasoning tokens on code tasks at lower compression ratios. This robust cross-domain performance strongly validates the generalization capability of our approach.

\begin{table}[t]
\centering
\caption{Ablation Study Variant Comparison}
\label{tab:ablation_color}
\begin{tabular}{ccc}
\toprule
\textbf{Model Variant} & \textbf{Conditional} & \textbf{Trained RM} \\
\midrule
Base           & $\times$            & $\times$          \\
+ Conditional          & $\checkmark$        & $\times$           \\
+ RM-Tuned         & $\times$            & $\checkmark$        \\
Proposed (CTS)        & $\checkmark$        & $\checkmark$       \\
\bottomrule
\end{tabular}
\end{table}

Our ablation study aims to verify: 1) The effectiveness of using a well-trained reference model for selecting valuable tokens, and 2) The effectiveness of conditional token importance assessment compared to unconditional methods.
For the unconditional token perplexity assessment, we follow the framework established in \citet{jiang2023llmlingua}.

We introduce the following variants of our method for the ablation study, as shown in Table \ref{tab:ablation_color}:
(1) \textbf{Base}: Using an standard RM to unconditionally predict token perplexity in CoT;
(2) \textbf{+ Conditional}: Using an standard RM to predict conditional token perplexity in CoT;
(3) \textbf{+ RM-Tuned}: Using a well-trained RM to unconditionally predict token perplexity in CoT;
(4) \textbf{Proposed (CTS)}: Using a well-trained RM to predict conditional token perplexity in CoT.

\begin{figure}
    \centering
    \includegraphics[width=0.9\linewidth]{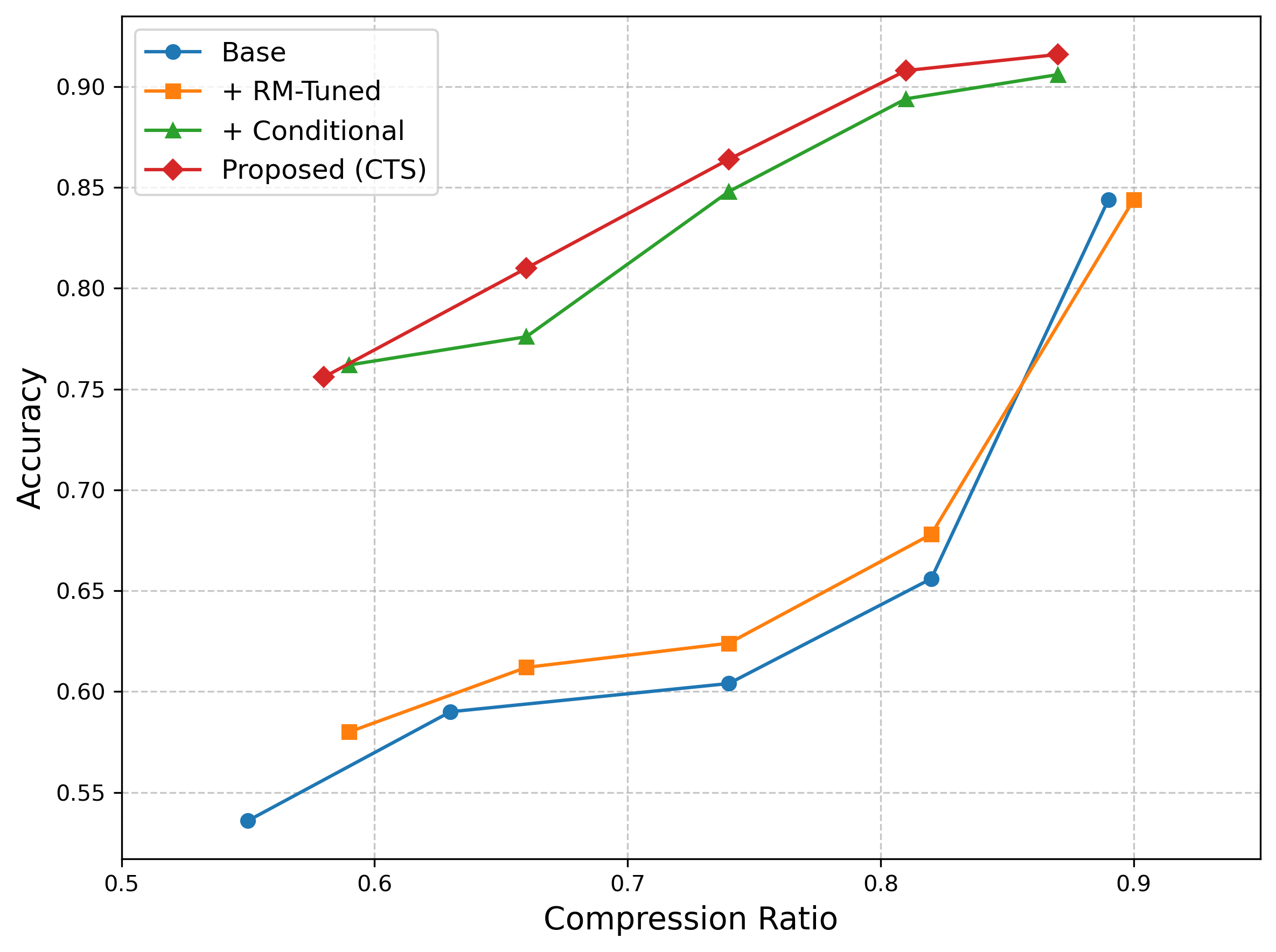}
    \caption{Ablation experiments of the Qwen2.5-14B-Instruct model on the MATH500 dataset under different compression ratios}
    \label{fig:ablation_acc}
\end{figure}

Conditional compression methods significantly enhance model performance. As shown in Figures \ref{fig:ablation_acc}, the top two curves—\textbf{Proposed (CTS)} and \textbf{+ Conditional}—leverage conditional token importance prediction, demonstrating their effectiveness in achieving higher accuracy. For both conditional and unconditional methods (\textbf{+ Conditional} and \textbf{Base}, respectively), the well-trained RM shifts the curves upward, further improving accuracy. This indicates that training the reference model provides some benefit in identifying important reasoning tokens. The accuracy improvement achieved by the reference model is relatively marginal compared to the conditional approach. Therefore, under circumstances with limited resources, employing \textbf{+ Conditional} alone can still yield satisfactory results.

Table 2 in appendix demonstrates that reasoning tokens do not differ substantially across variants, indicating that variants do not significantly improve inference efficiency. 
Interestingly, at certain compression ratios, while model accuracy improves, the number of inference tokens increases, resulting in reduced inference efficiency. This observation aligns with test time scaling results in \citet{zhang2025surveytesttimescalinglarge}, which indicate that model capability scales with inference length.


\section{Related Work}


\begin{figure}
    \centering
    \includegraphics[width=\linewidth]{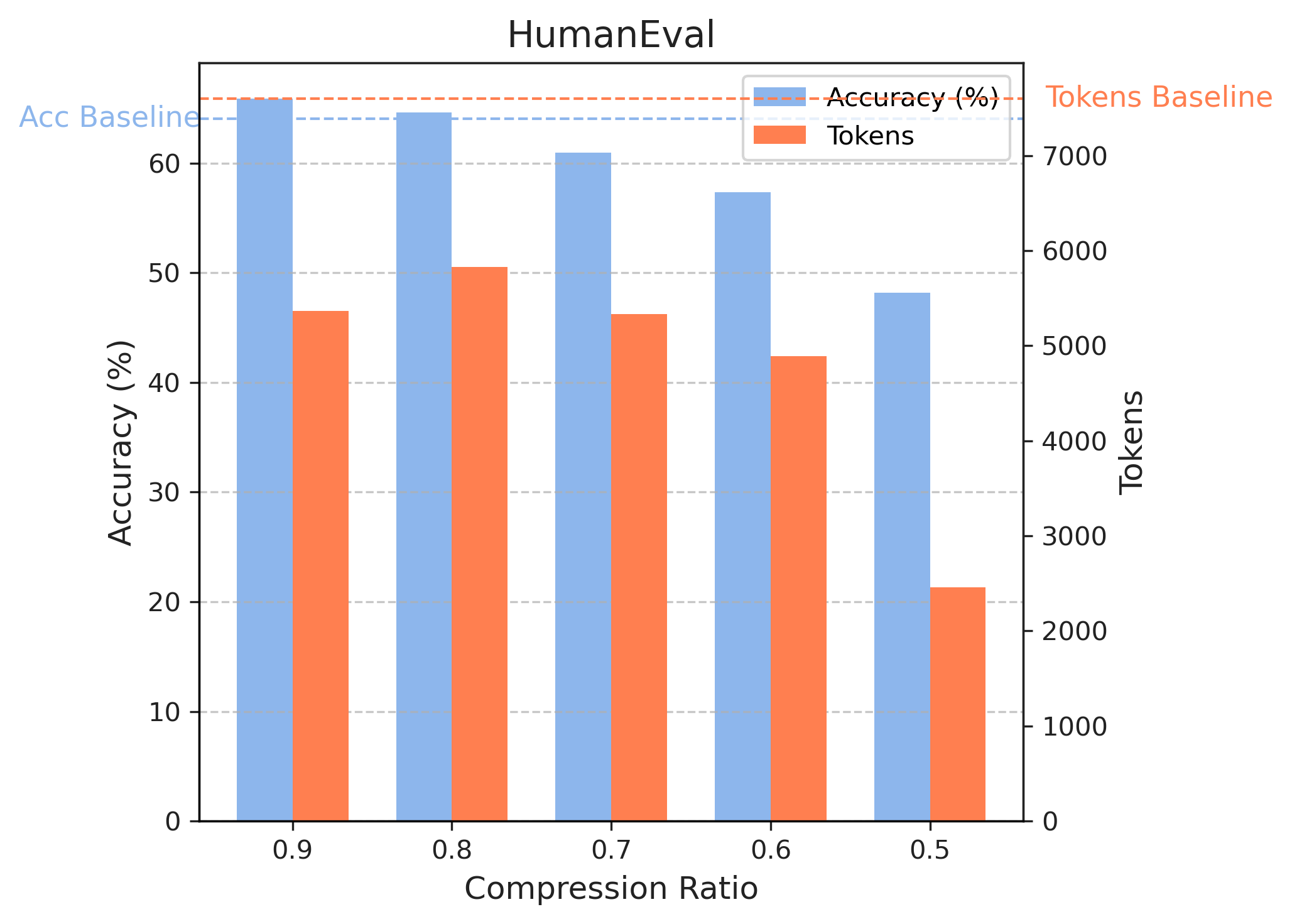}
    \caption{Generalization experiments of the CTS-14B model on the HumanEval datasets under different compression ratios}
    \label{fig:generalization}
\end{figure}

\subsection{Efficient Reasoning}
\citet{chen2024not, team2025kimi} demonstrated that in long CoT reasoning models, models generate overly detailed or unnecessarily elaborate reasoning steps, ultimately reducing their problem-solving efficiency.
Prompt-based Chain of Thought (CoT) methods \citep{wei2022chain, kojima2022large} guide models to think step-by-step, enhancing their problem-solving capabilities. 
Chain of Draft \citep{xu2025chain}, through prompting, retains essential formulas and numbers in the thought chain, maintaining performance while reducing inference costs. 
\citet{wu2025effectively, ma2025reasoningmodelseffectivethinking} implement thought intervention by incorporating first-person prompts in the model's thinking process to achieve instruction following and efficient reasoning.


\citet{kang2025c3ot} improves reasoning efficiency by compressing the reasoning process.
\citet{yu2025longshortchainofthoughtmixturesupervised} trains models by collecting variable-length CoT reasoning data.
\citet{munkhbat2025self, yeo2025demystifying, xia2025tokenskip} collect short CoT data by reducing redundant reasoning steps after full-length reasoning.

\subsection{Prompt Compression}
As language models increase in parameter scale, their capabilities also grow stronger, enabling many tasks to achieve better results merely by changing the model's input, such as RAG \citep{lewis2020retrieval} and few-shot learning \citep{wang2020generalizing}. As prompts become longer, resource consumption increases significantly. \citet{li2023compressing} proposed evaluating token importance by calculating token perplexity in context to compress the input.

Building on this foundation, \citet{jiang2023llmlingua} proposed a mixed coarse-grained and fine-grained compression method. \citet{jiang2023longllmlingua} extended this work by introducing a task-aware prompt compression method that incorporates conditional information from the query when calculating perplexity. \citet{pan2024llmlingua} transformed token compression into a binary classification problem, utilizing BERT as a classifier.

\section{Conclusion}
We introduce Conditional Token Selection , a novel method that employs a fine-tuned reference model to assess conditional perplexity differences in long CoT sequences, pinpointing the most reasoning-critical tokens for accurate answers. Through flexible compression ratios, CTS efficiently condenses CoT data while preserving reasoning quality and output efficiency.
Experimental results demonstrate CTS's effectiveness across multiple LLMs and tasks. Our method demonstrates strong generalization, even on out-of-domain (code) tasks. Ablation studies show that it remains effective without the reference model under resource constraints, further validating its robustness.




\bibliography{aaai2026}

\begin{thebibliography}{37}
\providecommand{\natexlab}[1]{#1}

\bibitem[{Austin et~al.(2021)Austin, Odena, Nye, Bosma, Michalewski, Dohan, Jiang, Cai, Terry, Le et~al.}]{austin2021program}
Austin, J.; Odena, A.; Nye, M.; Bosma, M.; Michalewski, H.; Dohan, D.; Jiang, E.; Cai, C.; Terry, M.; Le, Q.; et~al. 2021.
\newblock Program Synthesis with Large Language Models.
\newblock \emph{arXiv preprint arXiv:2108.07732}.

\bibitem[{Chen et~al.(2021)Chen, Tworek, Jun, Yuan, de~Oliveira~Pinto, Kaplan, Edwards, Burda, Joseph, Brockman, Ray, Puri, Krueger, Petrov, Khlaaf, Sastry, Mishkin, Chan, Gray, Ryder, Pavlov, Power, Kaiser, Bavarian, Winter, Tillet, Such, Cummings, Plappert, Chantzis, Barnes, Herbert-Voss, Guss, Nichol, Paino, Tezak, Tang, Babuschkin, Balaji, Jain, Saunders, Hesse, Carr, Leike, Achiam, Misra, Morikawa, Radford, Knight, Brundage, Murati, Mayer, Welinder, McGrew, Amodei, McCandlish, Sutskever, and Zaremba}]{chen2021evaluating}
Chen, M.; Tworek, J.; Jun, H.; Yuan, Q.; de~Oliveira~Pinto, H.~P.; Kaplan, J.; Edwards, H.; Burda, Y.; Joseph, N.; Brockman, G.; Ray, A.; Puri, R.; Krueger, G.; Petrov, M.; Khlaaf, H.; Sastry, G.; Mishkin, P.; Chan, B.; Gray, S.; Ryder, N.; Pavlov, M.; Power, A.; Kaiser, L.; Bavarian, M.; Winter, C.; Tillet, P.; Such, F.~P.; Cummings, D.; Plappert, M.; Chantzis, F.; Barnes, E.; Herbert-Voss, A.; Guss, W.~H.; Nichol, A.; Paino, A.; Tezak, N.; Tang, J.; Babuschkin, I.; Balaji, S.; Jain, S.; Saunders, W.; Hesse, C.; Carr, A.~N.; Leike, J.; Achiam, J.; Misra, V.; Morikawa, E.; Radford, A.; Knight, M.; Brundage, M.; Murati, M.; Mayer, K.; Welinder, P.; McGrew, B.; Amodei, D.; McCandlish, S.; Sutskever, I.; and Zaremba, W. 2021.
\newblock Evaluating Large Language Models Trained on Code.
\newblock arXiv:2107.03374.

\bibitem[{Chen et~al.(2025)Chen, Xu, Liang, He, Pang, Yu, Song, Liu, Zhou, Zhang, Wang, Tu, Mi, and Yu}]{chen2024not}
Chen, X.; Xu, J.; Liang, T.; He, Z.; Pang, J.; Yu, D.; Song, L.; Liu, Q.; Zhou, M.; Zhang, Z.; Wang, R.; Tu, Z.; Mi, H.; and Yu, D. 2025.
\newblock Do NOT Think That Much for 2+3=? On the Overthinking of o1-Like LLMs.
\newblock arXiv:2412.21187.

\bibitem[{Face(2025)}]{openr1}
Face, H. 2025.
\newblock Open R1: A fully open reproduction of DeepSeek-R1.

\bibitem[{Grattafiori, Dubey, and et~al.(2024)}]{grattafiori2024llama3herdmodels}
Grattafiori, A.; Dubey, A.; and et~al., A.~J. 2024.
\newblock The Llama 3 Herd of Models.
\newblock arXiv:2407.21783.

\bibitem[{Guo et~al.(2025)Guo, Yang, Zhang, Song, Zhang, Xu, Zhu, Ma, Wang, Bi et~al.}]{guo2025deepseek}
Guo, D.; Yang, D.; Zhang, H.; Song, J.; Zhang, R.; Xu, R.; Zhu, Q.; Ma, S.; Wang, P.; Bi, X.; et~al. 2025.
\newblock Deepseek-r1: Incentivizing reasoning capability in llms via reinforcement learning.
\newblock \emph{arXiv preprint arXiv:2501.12948}.

\bibitem[{Hendrycks et~al.(2021{\natexlab{a}})Hendrycks, Burns, Basart, Zou, Mazeika, Song, and Steinhardt}]{hendryckstest2021}
Hendrycks, D.; Burns, C.; Basart, S.; Zou, A.; Mazeika, M.; Song, D.; and Steinhardt, J. 2021{\natexlab{a}}.
\newblock Measuring Massive Multitask Language Understanding.
\newblock \emph{Proceedings of the International Conference on Learning Representations (ICLR)}.

\bibitem[{Hendrycks et~al.(2021{\natexlab{b}})Hendrycks, Burns, Kadavath, Arora, Basart, Tang, Song, and Steinhardt}]{hendrycks2021measuring}
Hendrycks, D.; Burns, C.; Kadavath, S.; Arora, A.; Basart, S.; Tang, E.; Song, D.; and Steinhardt, J. 2021{\natexlab{b}}.
\newblock Measuring mathematical problem solving with the math dataset.
\newblock \emph{arXiv preprint arXiv:2103.03874}.

\bibitem[{Jaech et~al.(2024)Jaech, Kalai, Lerer, Richardson, El-Kishky, Low, Helyar, Madry, Beutel, Carney et~al.}]{jaech2024openai}
Jaech, A.; Kalai, A.; Lerer, A.; Richardson, A.; El-Kishky, A.; Low, A.; Helyar, A.; Madry, A.; Beutel, A.; Carney, A.; et~al. 2024.
\newblock Openai o1 system card.
\newblock \emph{arXiv preprint arXiv:2412.16720}.

\bibitem[{Jiang et~al.(2023)Jiang, Wu, Lin, Yang, and Qiu}]{jiang2023llmlingua}
Jiang, H.; Wu, Q.; Lin, C.-Y.; Yang, Y.; and Qiu, L. 2023.
\newblock {LLML}ingua: Compressing Prompts for Accelerated Inference of Large Language Models.
\newblock In Bouamor, H.; Pino, J.; and Bali, K., eds., \emph{Proceedings of the 2023 Conference on Empirical Methods in Natural Language Processing}, 13358--13376. Singapore: Association for Computational Linguistics.

\bibitem[{Jiang et~al.(2024)Jiang, Wu, Luo, Li, Lin, Yang, and Qiu}]{jiang2023longllmlingua}
Jiang, H.; Wu, Q.; Luo, X.; Li, D.; Lin, C.-Y.; Yang, Y.; and Qiu, L. 2024.
\newblock {L}ong{LLML}ingua: Accelerating and Enhancing {LLM}s in Long Context Scenarios via Prompt Compression.
\newblock In Ku, L.-W.; Martins, A.; and Srikumar, V., eds., \emph{Proceedings of the 62nd Annual Meeting of the Association for Computational Linguistics (Volume 1: Long Papers)}, 1658--1677. Bangkok, Thailand: Association for Computational Linguistics.

\bibitem[{Kang et~al.(2025)Kang, Sun, Chen, and Zou}]{kang2025c3ot}
Kang, Y.; Sun, X.; Chen, L.; and Zou, W. 2025.
\newblock C3ot: Generating shorter chain-of-thought without compromising effectiveness.
\newblock In \emph{Proceedings of the AAAI Conference on Artificial Intelligence}, volume~39, 24312--24320.

\bibitem[{Kaplan et~al.(2020)Kaplan, McCandlish, Henighan, Brown, Chess, Child, Gray, Radford, Wu, and Amodei}]{kaplan2020scalinglawsneurallanguage}
Kaplan, J.; McCandlish, S.; Henighan, T.; Brown, T.~B.; Chess, B.; Child, R.; Gray, S.; Radford, A.; Wu, J.; and Amodei, D. 2020.
\newblock Scaling Laws for Neural Language Models.
\newblock arXiv:2001.08361.

\bibitem[{Kojima et~al.(2022)Kojima, Gu, Reid, Matsuo, and Iwasawa}]{kojima2022large}
Kojima, T.; Gu, S.~S.; Reid, M.; Matsuo, Y.; and Iwasawa, Y. 2022.
\newblock Large language models are zero-shot reasoners.
\newblock \emph{Advances in neural information processing systems}, 35: 22199--22213.

\bibitem[{Lewis et~al.(2020)Lewis, Perez, Piktus, Petroni, Karpukhin, Goyal, K{\"u}ttler, Lewis, Yih, Rockt{\"a}schel et~al.}]{lewis2020retrieval}
Lewis, P.; Perez, E.; Piktus, A.; Petroni, F.; Karpukhin, V.; Goyal, N.; K{\"u}ttler, H.; Lewis, M.; Yih, W.-t.; Rockt{\"a}schel, T.; et~al. 2020.
\newblock Retrieval-augmented generation for knowledge-intensive nlp tasks.
\newblock \emph{Advances in neural information processing systems}, 33: 9459--9474.

\bibitem[{Li et~al.(2023)Li, Dong, Lin, and Guerin}]{li2023compressing}
Li, Y.; Dong, B.; Lin, C.; and Guerin, F. 2023.
\newblock Compressing context to enhance inference efficiency of large language models.
\newblock \emph{arXiv preprint arXiv:2310.06201}.

\bibitem[{Ma et~al.(2025)Ma, He, Snell, Griggs, Min, and Zaharia}]{ma2025reasoningmodelseffectivethinking}
Ma, W.; He, J.; Snell, C.; Griggs, T.; Min, S.; and Zaharia, M. 2025.
\newblock Reasoning Models Can Be Effective Without Thinking.
\newblock arXiv:2504.09858.

\bibitem[{{Mathematical Association of America}(2024)}]{AoPS_AIME_2024}
{Mathematical Association of America}. 2024.
\newblock AIME.
\newblock URL \url{https://artofproblemsolving.com/wiki/index.php/AIME_Problems_and_Solutions/}.

\bibitem[{Moshkov et~al.(2025)Moshkov, Hanley, Sorokin, Toshniwal, Henkel, Schifferer, Du, and Gitman}]{moshkov2025aimo}
Moshkov, I.; Hanley, D.; Sorokin, I.; Toshniwal, S.; Henkel, C.; Schifferer, B.; Du, W.; and Gitman, I. 2025.
\newblock AIMO-2 Winning Solution: Building State-of-the-Art Mathematical Reasoning Models with OpenMathReasoning dataset.
\newblock \emph{arXiv preprint arXiv:2504.16891}.

\bibitem[{Muennighoff et~al.(2025)Muennighoff, Yang, Shi, Li, Fei-Fei, Hajishirzi, Zettlemoyer, Liang, Candès, and Hashimoto}]{muennighoff2025s1simpletesttimescaling}
Muennighoff, N.; Yang, Z.; Shi, W.; Li, X.~L.; Fei-Fei, L.; Hajishirzi, H.; Zettlemoyer, L.; Liang, P.; Candès, E.; and Hashimoto, T. 2025.
\newblock s1: Simple test-time scaling.
\newblock arXiv:2501.19393.

\bibitem[{Munkhbat et~al.(2025)Munkhbat, Ho, Kim, Yang, Kim, and Yun}]{munkhbat2025self}
Munkhbat, T.; Ho, N.; Kim, S.~H.; Yang, Y.; Kim, Y.; and Yun, S.-Y. 2025.
\newblock Self-Training Elicits Concise Reasoning in Large Language Models.
\newblock arXiv:2502.20122.

\bibitem[{Pan et~al.(2024)Pan, Wu, Jiang, Xia, and et~al.}]{pan2024llmlingua}
Pan, Z.; Wu, Q.; Jiang, H.; Xia, M.; and et~al. 2024.
\newblock {LLML}ingua-2: Data Distillation for Efficient and Faithful Task-Agnostic Prompt Compression.
\newblock In Ku, L.-W.; Martins, A.; and Srikumar, V., eds., \emph{Findings of the Association for Computational Linguistics: ACL 2024}, 963--981. Bangkok, Thailand: Association for Computational Linguistics.

\bibitem[{Qwen et~al.(2025)Qwen, :, Yang, Yang, Zhang, Hui, Zheng, Yu, Li, Liu, Huang, Wei, Lin, Yang, Tu, Zhang, Yang, Yang, Zhou, Lin, Dang, Lu, Bao, Yang, Yu, Li, Xue, Zhang, Zhu, Men, Lin, Li, Tang, Xia, Ren, Ren, Fan, Su, Zhang, Wan, Liu, Cui, Zhang, and Qiu}]{yang2024qwen2}
Qwen; :; Yang, A.; Yang, B.; Zhang, B.; Hui, B.; Zheng, B.; Yu, B.; Li, C.; Liu, D.; Huang, F.; Wei, H.; Lin, H.; Yang, J.; Tu, J.; Zhang, J.; Yang, J.; Yang, J.; Zhou, J.; Lin, J.; Dang, K.; Lu, K.; Bao, K.; Yang, K.; Yu, L.; Li, M.; Xue, M.; Zhang, P.; Zhu, Q.; Men, R.; Lin, R.; Li, T.; Tang, T.; Xia, T.; Ren, X.; Ren, X.; Fan, Y.; Su, Y.; Zhang, Y.; Wan, Y.; Liu, Y.; Cui, Z.; Zhang, Z.; and Qiu, Z. 2025.
\newblock Qwen2.5 Technical Report.
\newblock arXiv:2412.15115.

\bibitem[{Rein et~al.(2024)Rein, Hou, Stickland, Petty, Pang, Dirani, Michael, and Bowman}]{rein2024gpqa}
Rein, D.; Hou, B.~L.; Stickland, A.~C.; Petty, J.; Pang, R.~Y.; Dirani, J.; Michael, J.; and Bowman, S.~R. 2024.
\newblock Gpqa: A graduate-level google-proof q\&a benchmark.
\newblock In \emph{First Conference on Language Modeling}.

\bibitem[{Sui et~al.(2025)Sui, Chuang, Wang, Zhang, Zhang, Yuan, Liu, Wen, Chen, Hu et~al.}]{sui2025stop}
Sui, Y.; Chuang, Y.-N.; Wang, G.; Zhang, J.; Zhang, T.; Yuan, J.; Liu, H.; Wen, A.; Chen, H.; Hu, X.; et~al. 2025.
\newblock Stop overthinking: A survey on efficient reasoning for large language models.
\newblock \emph{arXiv preprint arXiv:2503.16419}.

\bibitem[{Team et~al.(2025)Team, Du, Gao, Xing, Jiang, and et~al.}]{team2025kimi}
Team, K.; Du, A.; Gao, B.; Xing, B.; Jiang, C.; and et~al. 2025.
\newblock Kimi k1.5: Scaling Reinforcement Learning with LLMs.
\newblock arXiv:2501.12599.

\bibitem[{Team(2025)}]{sky_t1_2025}
Team, N. 2025.
\newblock Sky-T1: Train your own O1 preview model within \$450.
\newblock https://novasky-ai.github.io/posts/sky-t1.
\newblock Accessed: 2025-01-09.

\bibitem[{Tian et~al.(2025)Tian, Zhao, Wang, Chen, Ji, Peng, Zhao, and Li}]{tian2025think}
Tian, X.; Zhao, S.; Wang, H.; Chen, S.; Ji, Y.; Peng, Y.; Zhao, H.; and Li, X. 2025.
\newblock Think twice: Enhancing llm reasoning by scaling multi-round test-time thinking.
\newblock \emph{arXiv preprint arXiv:2503.19855}.

\bibitem[{Wang et~al.(2020)Wang, Yao, Kwok, and Ni}]{wang2020generalizing}
Wang, Y.; Yao, Q.; Kwok, J.~T.; and Ni, L.~M. 2020.
\newblock Generalizing from a few examples: A survey on few-shot learning.
\newblock \emph{ACM computing surveys (csur)}, 53(3): 1--34.

\bibitem[{Wei et~al.(2022)Wei, Wang, Schuurmans, Bosma, Ichter, Xia, Chi, Le, and Zhou}]{wei2022chain}
Wei, J.; Wang, X.; Schuurmans, D.; Bosma, M.; Ichter, B.; Xia, F.; Chi, E.~H.; Le, Q.~V.; and Zhou, D. 2022.
\newblock Chain-of-Thought Prompting Elicits Reasoning in Large Language Models.
\newblock In \emph{Advances in Neural Information Processing Systems}.

\bibitem[{Wu et~al.(2025)Wu, Xiang, Wang, and Mittal}]{wu2025effectively}
Wu, T.; Xiang, C.; Wang, J.~T.; and Mittal, P. 2025.
\newblock Effectively Controlling Reasoning Models through Thinking Intervention.
\newblock arXiv:2503.24370.

\bibitem[{Xia et~al.(2025)Xia, Li, Leong, Wang, and Li}]{xia2025tokenskip}
Xia, H.; Li, Y.; Leong, C.~T.; Wang, W.; and Li, W. 2025.
\newblock Tokenskip: Controllable chain-of-thought compression in llms.
\newblock \emph{arXiv preprint arXiv:2502.12067}.

\bibitem[{Xu et~al.(2025)Xu, Xie, Zhao, and He}]{xu2025chain}
Xu, S.; Xie, W.; Zhao, L.; and He, P. 2025.
\newblock Chain of draft: Thinking faster by writing less.
\newblock \emph{arXiv preprint arXiv:2502.18600}.

\bibitem[{Yeo et~al.(2025)Yeo, Tong, Niu, Neubig, and Yue}]{yeo2025demystifying}
Yeo, E.; Tong, Y.; Niu, M.; Neubig, G.; and Yue, X. 2025.
\newblock Demystifying Long Chain-of-Thought Reasoning in LLMs.
\newblock arXiv:2502.03373.

\bibitem[{Yu et~al.(2025)Yu, Yuan, Wei, Wang, Qi, and Chen}]{yu2025longshortchainofthoughtmixturesupervised}
Yu, B.; Yuan, H.; Wei, Y.; Wang, B.; Qi, W.; and Chen, K. 2025.
\newblock Long-Short Chain-of-Thought Mixture Supervised Fine-Tuning Eliciting Efficient Reasoning in Large Language Models.
\newblock arXiv:2505.03469.

\bibitem[{Zhang et~al.(2025)Zhang, Lyu, Sun, Wang, Zhang, Hua, Wu, Guo, Wang, Muennighoff, King, Liu, and Ma}]{zhang2025surveytesttimescalinglarge}
Zhang, Q.; Lyu, F.; Sun, Z.; Wang, L.; Zhang, W.; Hua, W.; Wu, H.; Guo, Z.; Wang, Y.; Muennighoff, N.; King, I.; Liu, X.; and Ma, C. 2025.
\newblock A Survey on Test-Time Scaling in Large Language Models: What, How, Where, and How Well?
\newblock arXiv:2503.24235.

\bibitem[{Zhao et~al.(2025)Zhao, Wang, Peng, Zhao, Tian, Chen, Ji, and Li}]{zhao202514millionopensourcedistilled}
Zhao, H.; Wang, H.; Peng, Y.; Zhao, S.; Tian, X.; Chen, S.; Ji, Y.; and Li, X. 2025.
\newblock 1.4 Million Open-Source Distilled Reasoning Dataset to Empower Large Language Model Training.
\newblock arXiv:2503.19633.

\end{thebibliography}



\appendix



\newpage

\section{Prompt Template}

\subsection{Prompt Template For Training}

\begin{tcolorbox}[title=Supervised Fine Tune,label=box:sft]
Given the following problem, solve it step by step. \\

QUESTION: \{question\} \\
 
$<$think$>$
\{thought\_process\}
$<$/think$>$ \\

\{Final Answer\}
\end{tcolorbox}

\subsection{Prompt Template for Obtaining High-Quality Corpus}

\begin{tcolorbox}
Compress the given reasoning steps to short expressions, and such that you (Deepseek) can understand reasoning and reconstruct it as close as possible to the original. \\
Unlike the usual text compression, I need you to comply with the 5 conditions below: \\

1. You can ONLY remove unimportant words. \\
2. Do not reorder the original words.\\
3. Do not change the original words.\\
4. Do not use abbreviations or emojis.\\
5. Do not add new words or symbols.\\

Compress the origin aggressively by removing words only. Compress the origin as short as you can, while retaining as much
information as possible.
If you understand, please compress the following reasoning steps: \\

\{reasoning\_steps\}\\

The compressed reasoning steps are:\\

\end{tcolorbox}

\section{Case Study}
\label{sec:case_study}

As shown in Table \ref{tab:case_study}, the compressed model does learn to omit unimportant tokens, as indicated by the red text in the responses. However, TokenSkip may remove critical tokens required for the model to properly conclude its reasoning process. This can cause the model to enter a loop when repeatedly failing to generate a valid solution, as demonstrated by the blue text in the responses.

\begin{figure}[htbp]
    \centering
    \includegraphics[width=\linewidth]{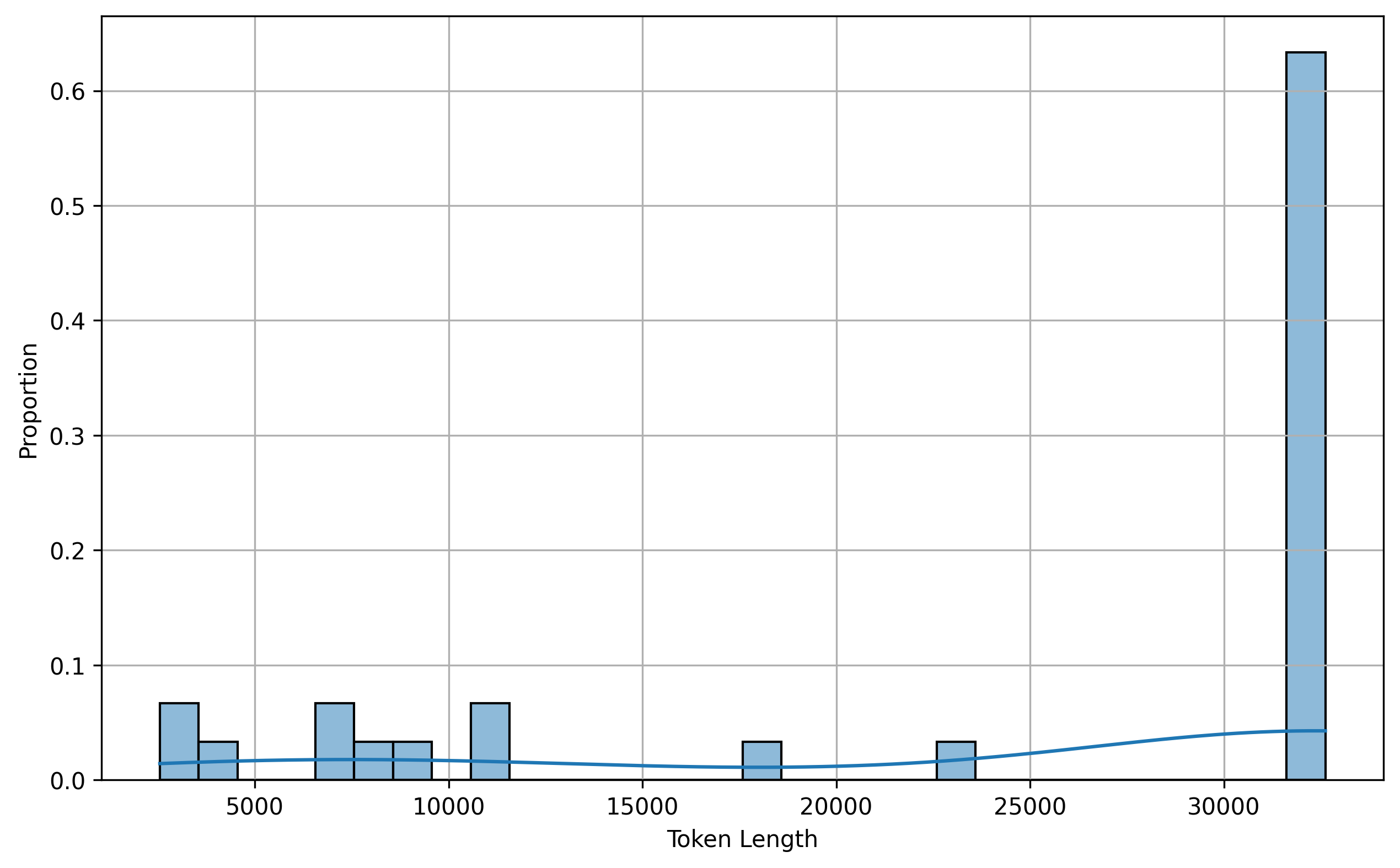}
    \caption{Inference token length distribution of Qwen2.5-7B-Instruct under TokenSkip with a compression ratio of 0.9 on AIME24 dataset.}
    \label{fig:case1}
\end{figure}

\begin{figure}[htbp]
    \centering
    \includegraphics[width=\linewidth]{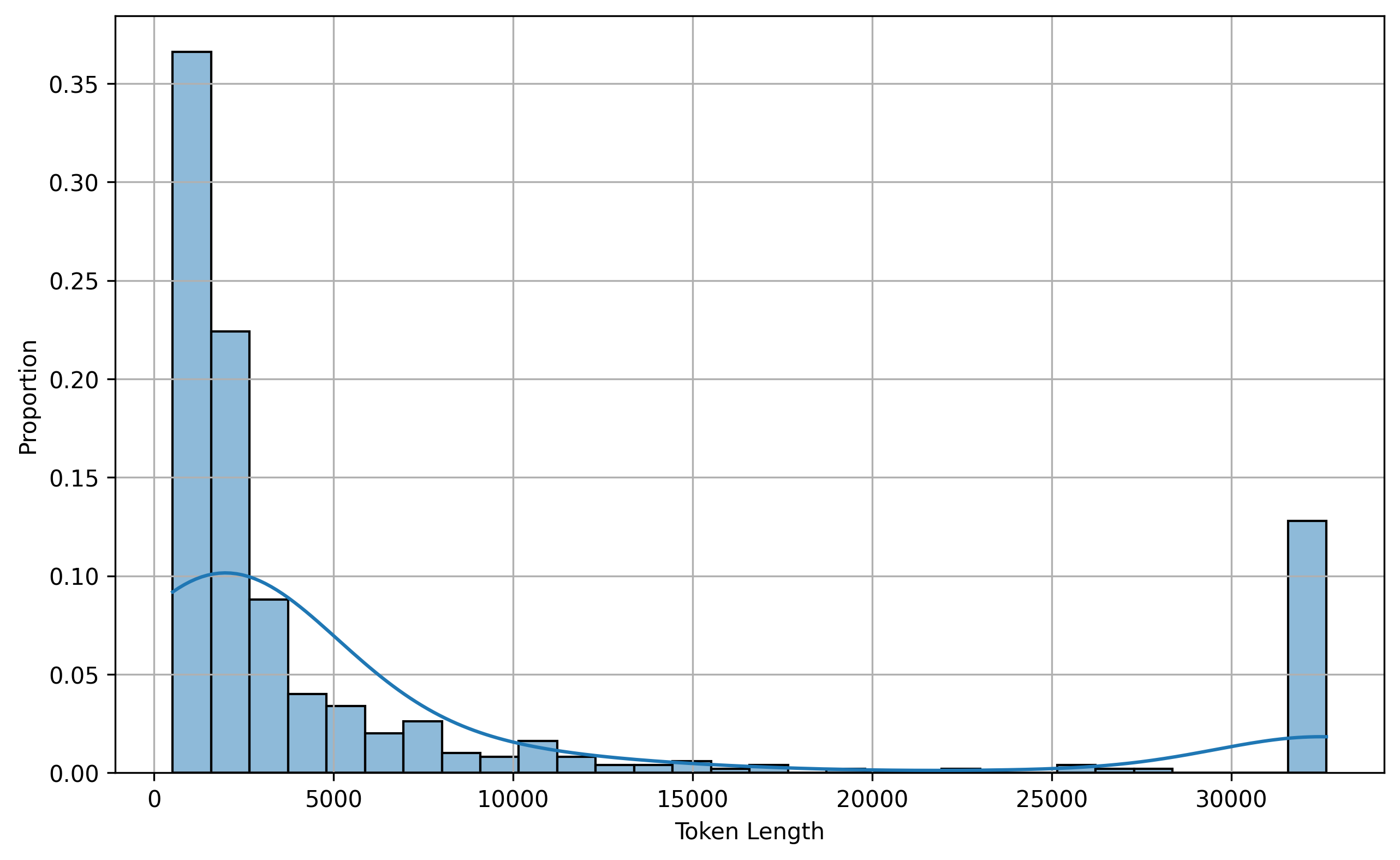}
    \caption{Inference token length distribution of Qwen2.5-7B-Instruct under TokenSkip with a compression ratio of 0.9 on MATH500 dataset.}
    \label{fig:case2}
\end{figure}

\begin{figure}[htbp]
    \centering
    \includegraphics[width=\linewidth]{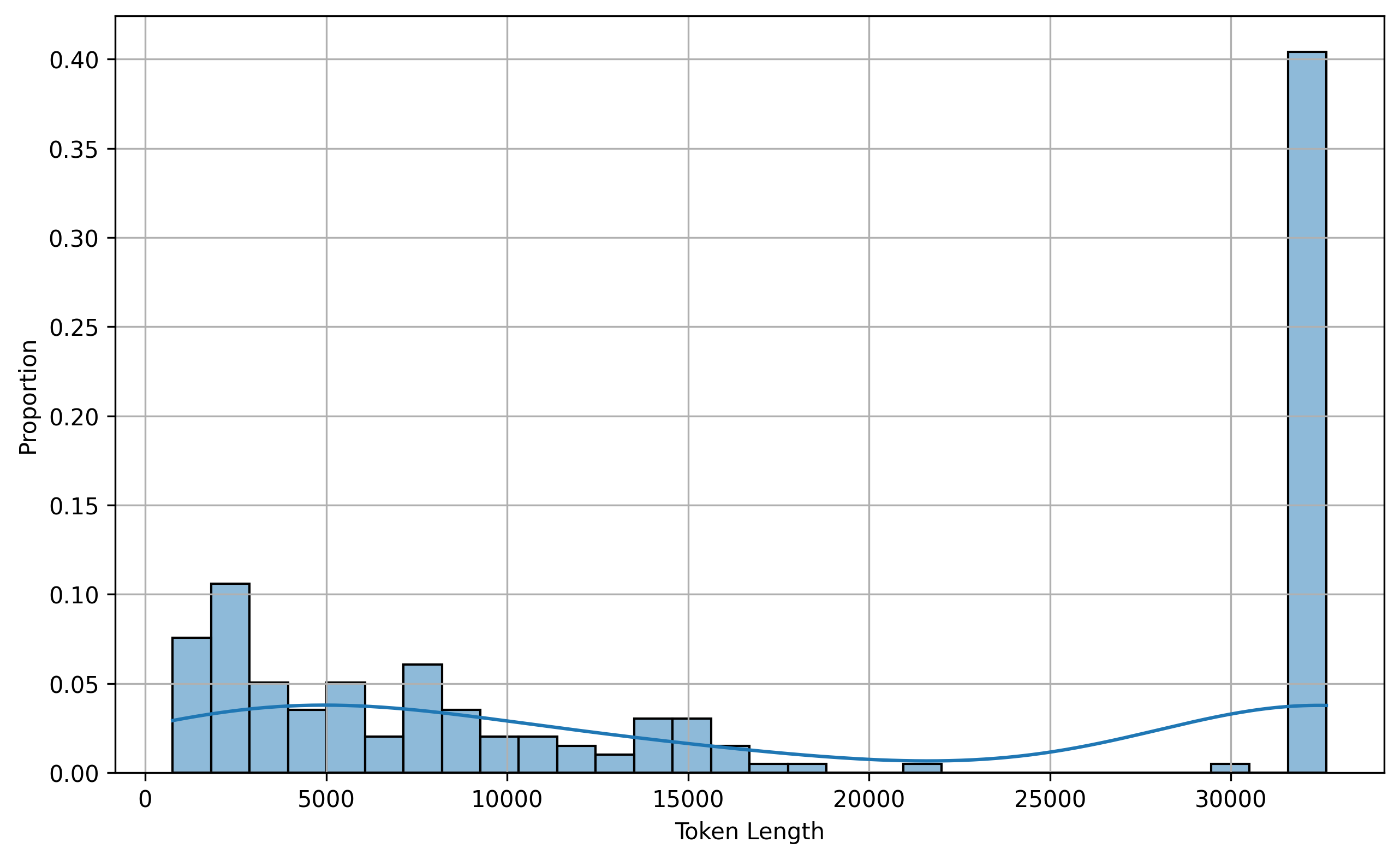}
    \caption{Inference token length distribution of Qwen2.5-7B-Instruct under TokenSkip with a compression ratio of 0.9 on GPQA-Diamond dataset.}
    \label{fig:case3}
\end{figure}

Furthermore, as shown in Figure \ref{fig:case1}, \ref{fig:case2} and \ref{fig:case3}, by examining the distribution of output token lengths across different benchmarks, we observe that most responses reach the maximum output length. This suggests that models compressed via TokenSkip lose the ability to properly terminate reasoning. Consequently, when faced with challenging problems, they are more prone to entering repetitive loops.

\begin{table*}[htbp]
\centering
\caption{A case of Qwen2.5-7B-Instruct under TokenSkip with a compression ratio of 0.9 on AIME24 dataset.}
\begin{tabular}{l}
\toprule
Question: Let $\omega\neq 1$ be a 13th root of unity. Find the remainder when
$\prod_{k=0}^{12}(2-2\omega^k+\omega^{2k})$
is divided by 1000. \\
\midrule
Response: $<$think$> $\\
\textcolor{red}{Okay, let's see. need} to find remainder when product from $k=0$ to 12 of $(2 - 2\omega^k + \omega^{2k})$ divided by 1000, \\
where $\omega$ is 13th root of unity and $\omega \neq 1$., sounds complicated, but maybe I can break it down. \\ 
First, since $\omega$ is a primitive 13th root of unity, means $\omega^13 = 1$, and $\omega^k \neq 1$ for $k$ from 1 to 12. \\
product is over $k$ from 0 to 12, includes $\omega^0$, is 1. check each term in product. For each $k$, term is $(2 - 2\omega^k + \omega^{2k})$. \\
see if can simplify this expression. Maybe factor it or relate it to some polynomial? \\
Alternatively, since $\omega$ is a root of unity, maybe way to express this product in terms of cyclotomic polynomials? \\
cyclotomic polynomials evaluated at $\omega$ might relate to products over roots. \\
\textcolor{red}{not sure. think step by step}. , note $\omega^{13} = 1$, so $\omega^{2k} = (\omega^k)^2$. \\
\textcolor{red}{Maybe can let} $x = \omega^k$, so each term becomes $2 - 2x + x^2$. \\
Then product becomes product from $x$ being 13th roots of unity (excluding $x=1$) of $(2 - 2x + x^2)$. \\
, no., in original product, each term is $(2 - 2\omega^k + \omega^{2k})$, which is $2 - 2\omega^k + (\omega^k)^2$. if let $x = \omega^k$, each term is $2 - 2x + x^2$. \\
product is product over all $x$, $x$ is 13th root of unity except $x=1$. \\
maybe can consider polynomial $P(x) = 2 - 2x + x^2$, and product is product of \\
$P(x)$ over all $x$ being 13th roots of unity except $x=1$.
how does that help? \\
Alternatively, maybe can factor $P(x) = x^2 - 2x + 2$. \\
\ldots \ldots \\
Since $\omega$ is a primitive 13th root of unity, modulo 8, $\omega$ is a root of unity not an integer, can't directly compute $\omega$.\\
, $\omega^{13} \equiv 1 mod 8$, $\omega^{13} - 1 \equiv 0 mod 8$. $\omega$ satisfies $x^{13} - 1 \equiv 0 mod 8$. $x^{13} - 1$ factors as $(x - 1)(x^{12} + x^{11} +. + x + 1) mod 8.$ \\
Since $\omega \neq 1$, it satisfies $x^{12} + x^{11} +. + x + 1 \equiv 0 mod 8$. \\
, $\omega^{12} + \omega^{11} +. + \omega + 1 \equiv 0 mod 8$., $\omega^{12} \equiv - (\omega^{11} +. + \omega + 1) mod 8$. \\
, $\omega^2$ is also a root of$ x^{12} + x^{11} +. + x + 1 \equiv 0 mod 8$,$ \omega^{24} + \omega^{22} +. + \omega^2 + 1 \equiv 0 mod 8$. \\
, compute$ A = 2 - 2\omega + \omega^2$. compute$ A + 2\omega - 2$. $A + 2\omega - 2 = (2 - 2\omega + \omega^2) + 2\omega - 2 = \omega^2 mod 8$., \\
$A + 2\omega - 2 = \omega^2 mod 8$.
, not sure. compute $A + 2\omega - 2$: \\
 $A + 2\omega - 2 = (2 - 2\omega + \omega^2) + 2\omega - 2 = \omega^2 mod 8., A + 2\omega - 2 = \omega^2 mod 8. A = \omega^2 - 2\omega + 2 mod 8$. \\
, \textcolor{blue}{not helpful}. \\
, compute $A^2$:
 $A^2 = (2 - 2\omega + \omega^2)^2 = 4 - 8\omega + 4\omega^2 + 4\omega^3 - 4\omega^4 + \omega^4 - 4\omega^3 + 2\omega^5 + \omega^6$. \\
, messy. \\

compute $A^2 modulo 8$. \\
, since $\omega^13 \equiv 1 mod 8, \omega^{13} \equiv 1 mod 8, \omega^{13k + m} \equiv \omega^m mod 8$., $\omega^4 \equiv \omega^{4 mod 12} \equiv \omega^4 mod 8$., $\omega^3 \equiv \omega^{3 mod 12} \equiv \omega^3 mod 8.,$ \\
$\omega^2, \omega^5, etc., modulo 8$.
, compute $A^2 = (2 - 2\omega + \omega^2)^2$: \\
, expand it:
 $(2 - 2\omega + \omega^2)^2 = 4 - 8\omega + 4\omega^2 + (-8\omega + 4\omega^2) + 4\omega^3 + (-4\omega^2 + 2\omega^3) + \omega^4$. \\
, step by step: \\
 $(a + b + c)^2 = a^2 + b^2 + c^2 + 2ab + 2ac + 2bc
 a = 2, b = -2\omega, c = \omega^2$.
,
 $a^2 = 4$
 $b^2 = (-2\omega)^2 = 4\omega^2$ \\
$ c^2 = (\omega^2)^2 = \omega^4$
$ 2ab = 2*2*(-2\omega) = -8\omega$
$ 2ac = 2*2*\omega^2 = 4\omega^2$
 $2bc = 2*(-2\omega)*\omega^2 = -4\omega^3$ \\
, combining all terms: \\
 $4 + 4\omega^2 + \omega^4 -8\omega + 4\omega^2 -4\omega^3$
 Combine terms: \\
 $4 + (4\omega^2 + 4\omega^2) + (-8\omega) + (-4\omega^3) + \omega^4
 4 + 8\omega^2 -8\omega -4\omega^3 + \omega^4 mod 8$.
, $A^2 = 4 + 8\omega^2 -8\omega -4\omega^3 + \omega^4 mod 8$. \\
 Simplify modulo 8: \\
 $4 + 0\omega^2 -8\omega -4\omega^3 + \omega^4 mod 8$.
 $-8\omega mod 8 is 0\omega, and 4 + 0 -4\omega^3 + \omega^4 mod 8$.
, $A^2 \equiv 4 - 4\omega^3 + \omega^4 mod 8$. \\
, \textcolor{blue}{not helpful}. \\
\ldots \ldots \\

\textcolor{blue}{another approach}. \\
, $compute A = 2 - 2\omega + \omega^2 mod 8. compute A + 2\omega - 2$: \\
 $A + 2\omega - 2 = (2 - 2\omega + \omega^2) + 2\omega - 2 = \omega^2 mod 8.,$ \\
 $ A + 2\omega - 2 = \omega^2 mod 8., A = \omega^2 - 2\omega + 2 mod 8$. \\
, $A = \omega^2 - 2\omega + 2 mod 8$. \\
, compute $A^6 = (\omega^2 - 2\omega + 2)^6 mod 8$. \\
, compute $A = \omega^2 - 2\omega + 2. compute A + 2\omega - 2 = \omega^2 mod 8, A + 2\omega - 2 = \omega^2., A = \omega^2 - 2\omega + 2 mod 8$. \\
, compute $A + 2\omega - 2 = \omega^2 mod 8, A + 2\omega - 2 = \omega^2., A = \omega^2 - 2\omega + 2 mod 8$. \\
, compute $A + 2\omega - 2 = \omega^2 mod 8, A = \omega^2 - 2\omega + 2 mod 8$. \\
, compute $A + 2\omega - 2 = \omega^2 mod 8, A = \omega^2 - 2\omega + 2 mod 8$. \\
, compute $A + 2\omega - 2 = \omega^2 mod 8, A = \omega^2 - 2\omega + 2 mod 8$. \\
\ldots \ldots \\

\bottomrule
\end{tabular}
\label{tab:case_study}
\end{table*}

\newpage

\section{Additional Experimental Results}

\begin{table}[h] 
\centering
\caption{Inference tokens of the Qwen2.5-14B-Instruct model on the MATH500 dataset across various compression ratios in ablation experiments.}
\label{tab:ablation_tokens}
\renewcommand{\arraystretch}{1} 
\resizebox{0.85\linewidth}{!}{
\begin{tabular}{ccccc} 
\toprule
Ratio & Base & + RM & + Conditional & Proposed \\
\midrule
0.9 & 5023  & 5597  & 4563  & 5012  \\
0.8 & 2510  & 2992  & 5001  & 4703  \\
0.7 & 2511  & 2369  & 2883  & 3310  \\
0.6 & 2270  & 2167  & 1952  & 3787  \\
0.5 & 1998  & 2036  & 1787  & 2036  \\
\bottomrule
\end{tabular}
}
\end{table}

\begin{table}[h]
\centering
\caption{Experimental results of the Qwen2.5-Instruct series on code benchmarks.}
\label{tab:general}
\begin{subtable}[t]{0.48\textwidth}
\centering
\caption{HumanEval benchmark}
\resizebox{0.85\linewidth}{!}{
\begin{tabular}{ccccc}
\toprule
Scale & Methods & ActRatio & Accuracy & Tokens \\
\midrule
\multirow{6}{*}{7B} & Original & - & 51.2  & 10153  \\ 
\cline{2-5}
 & \multirow{5}{*}{CTS} & 0.87 & 59.8  & 8205  \\
 &  & 0.81 & 54.9  & 9585  \\
 &  & 0.74 & 48.2  & 7718  \\
 &  & 0.66 & 44.5  & 2949  \\
 &  & 0.58 & 43.9  & 2970  \\
\midrule
\multirow{6}{*}{14B} & Original & - & 64.0  & 7595  \\ 
\cline{2-5}
 & \multirow{5}{*}{CTS} & 0.87 & 65.9  & 5365  \\
 &  & 0.81 & 64.6  & 5826  \\
 &  & 0.74 & 61.0  & 5333  \\
 &  & 0.66 & 57.3  & 4890  \\
 &  & 0.58 & 48.2  & 2459 \\
\bottomrule
\end{tabular}
}
\end{subtable}
\hfill
\begin{subtable}[t]{0.48\textwidth}
\centering
\caption{MBPP benchmark}
\resizebox{0.85\linewidth}{!}{
\begin{tabular}{ccccc}
\toprule
Scale & Methods & ActRatio & Accuracy & Tokens \\
\midrule
\multirow{6}{*}{7B} & Original & - & 51.2  & 10153  \\ 
\cline{2-5}
 & \multirow{5}{*}{TokenSkip} & 0.87 & 59.8  & 8205  \\
 &  & 0.81 & 54.9  & 9585  \\
 &  & 0.74 & 48.2  & 7718  \\
 &  & 0.66 & 44.5  & 2949  \\
 &  & 0.58 & 43.9  & 2970  \\
\midrule
\multirow{6}{*}{14B} & Original & - & 64.0  & 7595  \\ 
\cline{2-5}
 & \multirow{5}{*}{TokenSkip} & 0.87 & 65.9  & 5365  \\
 &  & 0.81 & 64.6  & 5826  \\
 &  & 0.74 & 61.0  & 5333  \\
 &  & 0.66 & 57.3  & 4890  \\
 &  & 0.58 & 48.2  & 2459 \\
\bottomrule
\end{tabular}
}
\end{subtable}
\end{table}

Table \ref{tab:ablation_tokens} presents the number of inference tokens for different variants under various compression ratios in the ablation study conducted on the MATH500 dataset. Table \ref{tab:general} demonstrates the results of the generalization experiments for CTS.

Tables \ref{tab:7B_all} and \ref{tab:llama} present the experimental results of various approaches on the Qwen2.5-7B-Instruct and Llama-3.1-8B-Instruct models, respectively.

\newpage

\begin{table*}[h]
  \centering
    \caption{Experimental results of various compression methods on \textbf{Qwen2.5-7B-Instruct}, showing accuracy, average reasoning CoT tokens, and compression ratio (actual ratio).}
  \begin{tabular}{@{}lccccccc@{}}
    \toprule
    \multirow{2}{*}{Methods} & \multirow{2}{*}{Ratio (Actual)} & \multicolumn{2}{c}{MATH500} & \multicolumn{2}{c}{AIME24} & \multicolumn{2}{c}{GPQA Diamond} \\
    \cmidrule(lr){3-4} \cmidrule(lr){5-6} \cmidrule(lr){7-8}
    & & Accuracy $\uparrow$ & Tokens $\downarrow$ & Accuracy $\uparrow$ & Tokens $\downarrow$ & Accuracy $\uparrow$ & Tokens $\downarrow$ \\
    \midrule
    Original & 1.0 & $ 82.4$ & 7244 & $20$ & 24396 & $37.8$ & 17038 \\
    \midrule
    \multirow{5}{*}{LLMLingua} 
    & 0.9(0.89) & $72.6$ & $6804$ & $13.3$ & \textbf{23903} & 39.8 & \textbf{14470} \\
    & 0.8(0.81) & $58.6$ & \textbf{2969} & $6.7$ & \textbf{4194} & 34.3 &\textbf{4421} \\
    & 0.7(0.73) & $57.2$ & \textbf{2542} & $6.7$ & \textbf{3692} & 32.8 & \textbf{3236} \\
    & 0.6(0.62) & $55.0$ & \textbf{2178} & $3.3$ & \textbf{3084} & \textbf{33.3} & 3203 \\
    & 0.5(0.55) & $51.4$ & $2226$ & $3.3$ & $3603$ & 30.8 & \textbf{2462} \\
    \midrule
    \multirow{5}{*}{TokenSkip} 
    & 0.9(0.88) & 78.6 & 6997 & \textbf{23.3} & 24094 & 38.8 & 17263 \\
    & 0.8(0.80) & 72.8 & 8172 & 10.0 & 25223 & 39.3 & 18365 \\
    & 0.7(0.71) & 64.2 & 9984 & 6.6 & 27946 & 32.3 & 21219 \\
    & 0.6(0.62) & 54.6 & 11496 & 3.3 & 28802 & 26.2 & 21371 \\
    & 0.5(0.50) & 37.4 & 13595 & 0 & 29470 & 31.3 & 21012 \\
    \midrule
    \multirow{5}{*}{\textbf{CTS}}
    & 0.9(0.87) & \textbf{82.8} & \textbf{6497} & 20 & 24769 & \textbf{43.4} & 17272 \\
    & 0.8(0.81) & \textbf{81.2} & $6886$ & \textbf{23.3} & 27006 & \textbf{39.3} & 17961 \\
    & 0.7(0.74) & \textbf{78.0} & $5109$ & \textbf{13.3} & 15929 & \textbf{42.4} & 13937 \\
    & 0.6(0.66) & \textbf{70.8} & $2198$ & \textbf{10.0} & 3550 & 32.3 & \textbf{3055} \\
    & 0.5(0.58) & \textbf{70.6} & \textbf{2039} & \textbf{6.7} & \textbf{2993} & \textbf{32.8} & 3187 \\
    
    \bottomrule
  \end{tabular}
  \label{tab:7B_all} 
\end{table*}

\begin{table*}[h]
  \centering
    \caption{Experimental results of various compression methods on \textbf{Llama-3.1-8B-Instruct}, showing accuracy, average reasoning CoT tokens, and compression ratio (actual ratio).}
    \begin{tabular}{@{}lccccccc@{}}
    \toprule
    \multirow{2}{*}{Methods} & \multirow{2}{*}{Ratio (Actual)} & \multicolumn{2}{c}{MATH500} & \multicolumn{2}{c}{AIME24} & \multicolumn{2}{c}{GPQA Diamond} \\
    \cmidrule(lr){3-4} \cmidrule(lr){5-6} \cmidrule(lr){7-8}
    & & Accuracy $\uparrow$ & Tokens $\downarrow$ & Accuracy $\uparrow$ & Tokens $\downarrow$ & Accuracy $\uparrow$ & Tokens $\downarrow$ \\
    \midrule
    Original & 1.0 & 63.2 & 12538 & 0.033 & 31013 & 0.363 & 19069 \\
    \midrule
    \multirow{5}{*}{LLMLingua} 
    & 0.9(0.89) & 54.8 & \textbf{9811} & 0    & \textbf{27327} & 30.3 & \textbf{11727} \\
    & 0.8(0.81) & 39.6 & \textbf{3917} & 3.3  & \textbf{5609}  & 37.4 & \textbf{5471}  \\
    & 0.7(0.73) & 36.2 & \textbf{3134} & 0    & \textbf{3602}  & 25.3 & \textbf{3323}  \\
    & 0.6(0.62) & 39.0 & 2863 & 0    & 4334  & \textbf{32.8} & 3548  \\
    & 0.5(0.55) & 34.8 & 3172 & 0    & 4188  & \textbf{30.3} & 3438  \\
    \midrule
    \multirow{5}{*}{TokenSkip} 
    & 0.9(0.88) & 56.4 & 11985 & 0 & 27882 & \textbf{35.8} & 17846 \\
    & 0.8(0.80) & 51.2 & 13145 & 3.3 & 30443 & 30.3 & 17960 \\
    & 0.7(0.71) & 44.6 & 14354 & 3.3 & 34249 & 27.7 & 19265 \\
    & 0.6(0.62) & 32.4 & 15453 & 0 & 23319 & 29.7 & 20800 \\
    & 0.5(0.50) & 23.3 & 16013 & 0 & 22318 & 26.7 & 22010 \\
    \midrule
    \multirow{5}{*}{\textbf{CTS}}
    & 0.9(0.87) & \textbf{60.6} & 12047 & \textbf{3.3} & 29144 & 32.3 & 18503 \\
    & 0.8(0.81) & \textbf{58.4} & 12134 & \textbf{6.7} & 24906 & \textbf{40.4} & 18981 \\
    & 0.7(0.74) & \textbf{55.0} & 9987 & \textbf{3.3} & 25933 & \textbf{32.3} & 16571 \\
    & 0.6(0.66) & \textbf{50.8} & \textbf{2808} & 0 & \textbf{3781} & 26.7 & \textbf{3492} \\
    & 0.5(0.58) & \textbf{45.5} & \textbf{2625} & 0 & \textbf{3478} & 29.2 & \textbf{3080} \\
    
    \bottomrule
  \end{tabular}
  \label{tab:llama} 
\end{table*}

\newpage

\newpage

\section{Reference Model Training Cost Analysis}

The key parameters for estimating the FLOPs in reference (RM) computation are:
\begin{itemize}
    \item Model parameters: $7 \times 10^9$
    \item Total tokens processed: $8 \times 10^6$
    \item Training epochs: 3
    \item Sequence length: 4,096
\end{itemize}

The FLOPs for the training process are calculated as:
\begin{equation}
    \text{FLOPs}_{\text{train}} \approx 6 \times \text{Parameters} \times \text{Sequence Length} \times \text{Total Tokens}
\end{equation}
yielding a total computational cost of $4.13 \times 10^{18}$ FLOPs.

\subsection{Inference Efficiency Gains}
For our reference model \texttt{Qwen-7B-Instruct} using \texttt{CTS-14B} at compression ratio 0.7, we observe:
\begin{itemize}
    \item Average token reduction per inference: 5,300 (across MATH-500, AIME24, and GPQA Diamond)
\end{itemize}

Using the inference FLOPs estimation from \citet{kaplan2020scalinglawsneurallanguage}:
\begin{equation}
    \text{FLOPs}_{\text{inf}} \approx 2 \times \text{Parameters} \times \text{Sequence Length}
\end{equation}

\subsection{Cost-Benefit Analysis}
\begin{itemize}
    \item FLOPs saved per inference: $1.5 \times 10^{14}$
    \item Break-even point: 27,500 inference calls
\end{itemize}

\begin{equation}
    N_{\text{break-even}} = \frac{\text{FLOPs}_{\text{train}}}{\text{FLOPs}_{\text{inf}}} \approx 27,\!500
\end{equation}

\end{document}